\documentclass[10pt,twocolumn,letterpaper]{article}

\usepackage{cvpr}
\usepackage{times}
\usepackage{epsfig}
\usepackage{graphicx}
\usepackage{amsmath}
\usepackage{amssymb}

\usepackage{algorithmic}
\usepackage{algorithm}
\usepackage{subcaption}
\usepackage{rotating}
\usepackage{diagbox}
\usepackage{multirow}
\usepackage{balance}

\usepackage[pagebackref=true,breaklinks=true,letterpaper=true,colorlinks,bookmarks=false]{hyperref}

\cvprfinalcopy 

\def\cvprPaperID{3284} 

\ifcvprfinal\fi

\usepackage{float}
\usepackage{xcolor}
\usepackage{siunitx}

\usepackage{paralist}

\newcommand{\crossmodal}{cross-modal }

\newcommand\figref{Fig.~\ref}

\begin{document}

\title{Cross-modal Deep Variational Hand Pose Estimation}

\author{Adrian Spurr, Jie Song, Seonwook Park, Otmar Hilliges\\
ETH Zurich\\
{\tt\small \{spurra,jsong,spark,otmarh\}@inf.ethz.ch}
}

\maketitle
\thispagestyle{empty}

\begin{abstract}
The human hand moves in complex and high-dimensional ways, making estimation of 3D hand pose configurations from images alone a challenging task.
In this work we propose a method to learn a statistical hand model represented by a \crossmodal trained latent space via a generative deep neural network.
We derive an objective function from the variational lower bound of the VAE framework and jointly optimize the resulting \crossmodal KL-divergence and the posterior reconstruction objective, naturally admitting a training regime that leads to a coherent latent space across multiple modalities such as RGB images, 2D keypoint detections or 3D hand configurations. Additionally, it grants a straightforward way of using semi-supervision.
This latent space can be directly used to estimate 3D hand poses from RGB images, outperforming the state-of-the art in different settings. Furthermore, we show that our proposed method can be used without changes on depth images and performs comparably to specialized methods.
Finally, the model is fully generative and can synthesize consistent pairs of hand configurations across modalities. We evaluate our method on both RGB and depth datasets and analyze the latent space qualitatively.

\end{abstract}


\section{Introduction}
Hands are of central importance to humans in manipulating the physical world and in communicating with each other. Recovering the spatial configuration of hands from natural images therefore has many important applications in AR/VR, robotics, rehabilitation and HCI. Much work exists that tracks articulated hands in streams of depth images, or that estimates hand pose~\cite{oberweger2017,oberweger2015,tang2014,wan2016} from individual depth frames. However, estimating the full 3D hand pose from monocular RGB images only is a more challenging task due to the manual dexterity, symmetries and self-similarities of human hands as well as difficulties stemming from occlusions, varying lighting conditions and lack of accurate scale estimates. Compared to depth images the RGB case is less well studied.

\begin{figure}[t]
\begin{center}
 \includegraphics[width=1.0\linewidth, clip,trim={8mm 10mm 5mm 8mm}]{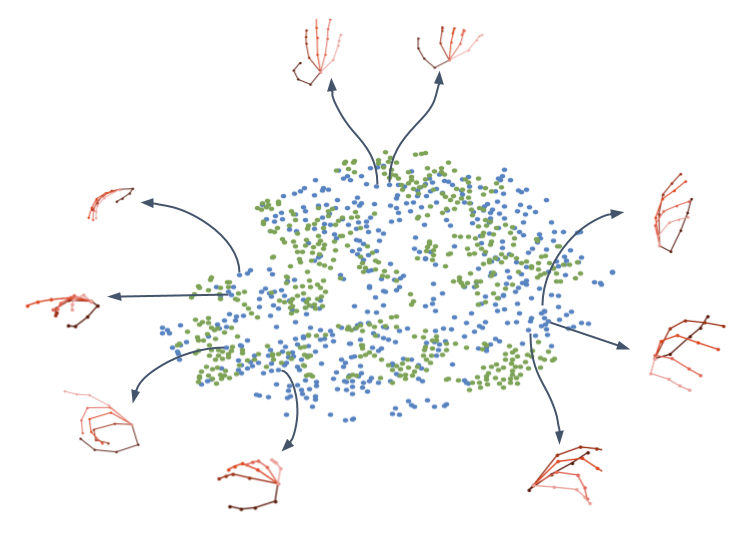}
\end{center}
\vspace*{-6mm}
   \caption{\textbf{Cross-modal latent space.} t-SNE visualization of 500 input samples of different modalities in the latent space. Embeddings of RGB images are shown in blue, embeddings of 3D joint configurations in green. Hand poses are decoded samples drawn from the latent space. Embedding does not cluster by modality, showing that there is a unified latent space. The posterior across different modalities can be estimated by sampling from this manifold.}
   \label{fig:teaser}

\end{figure}
Recent work relying solely on RGB images~\cite{zimmermann2017} proposes a deep learning architecture that decomposes the task into several substeps, demonstrating initial feasibility and providing a public dataset for comparison. The proposed architecture is specifically designed for the monocular case and splits the task into hand and 2D keypoint detection followed by a 2D-3D lifting step but incorporates no explicit hand model. Our work is also concerned with the estimation of 3D joint-angle configurations of human hands from RGB images but learns a cross-modal, statistical hand model. This is attained via learning of a latent representation that embeds sample points from multiple data sources such as 2D keypoints, images and 3D hand poses.
Samples from this latent space can then be reconstructed by \emph{independent} decoders to produce \emph{consistent} and physically plausible 2D or 3D joint predictions and even RGB images.

Findings from bio-mechanics suggest that while articulated hands have many degrees-of-freedom, only few are fully independently articulated~\cite{santello1998postural}. Therefore a sub-space of valid hand poses is supposed to exist and prior work on depth based hand tracking~\cite{tagliasacchi_sgp15} has successfully employed dimensionality reduction techniques to improve accuracy.

This idea has been recently revisited in the context of deep-learning, where Wan et al. \cite{wan2017} attempt to learn a manifold of hand poses via a combination of variational autoencoders (VAEs) and generative adversarial networks (GANs) for hand pose estimation from depth images. However, their approach is based on two separate manifolds, one for 3D hand joints (VAE) and one for depth-maps (GAN) and requires a mapping function between the two.

In this work we propose to learn a single, unified latent space via an extension of the VAE framework. We provide a derivation of the variational lower bound that permits training of a single latent space using multiple modalities, where similar input poses are embedded close to each other independent of the input modality. Fig. \ref{fig:teaser} visualizes this learned unified latent space for two modalities (RGB \& 3D). We focus on RGB images and hence test the architecture on different combinations of modalities where the goal is to produce 3D hand poses as output. At the same time, the VAE framework naturally allows to generate samples consistently in any modality.

We experimentally show that the proposed approach outperforms the state-of-the art method \cite{zimmermann2017} in direct RGB to 3D hand pose estimation, as well as in lifting from 2D detections to 3D on a challenging public dataset. Meantime, we note that given any input modality a mapping into the embedding space can be found and likewise hand configurations can be reconstructed in various modalities, thus the approach learns a many-to-many mapping. We demonstrate this capability via generation of novel hand pose configurations via sampling from the latent space and consistent reconstruction in different modalities (i.e., 3D joint positions and synthesized RGB images).
These could be potentially used in hybrid approaches for temporal tracking
or to generate additional training data. Furthermore, we explore the utility of the same architecture in the case of depth images and show that we are comparable
to state-of-art depth based methods \cite{oberweger2017,oberweger2015,wan2017} that employ specialized architectures.


\section{Related Work}
Capturing the 3D motion of human hands from images is a long standing problem in computer vision and related areas (cf. \cite{erol2007vision}). With the recent emergence of consumer grade RGB-D sensors and increased importance of AR and VR this problem has seen increased attention~\cite{sharp2015accurate,sun2015cascaded,tagliasacchi_sgp15,tang2014,tang2015opening,tang2013real,taylor2016efficient,wan2017,wan2016,zhang2016}. Generally speaking approaches can be categorized into tracking of articulated hand motion over time (e.g.,~\cite{oikonomidis2011efficient}) and per-frame classification~\cite{sun2015cascaded,tang2014,wan2017}. Furthermore, a number of hybrid methods exist that first leverage a discriminative model to initialize a hand pose estimate which is then refined and tracked via carefully designed energy functions to fit a hand model into the observed depth data~\cite{qian2014realtime,sharp2015accurate,taylor2016efficient,tzionas2016capturing,ye2016spatial}. Estimating hand pose from RGB images is more challenging.

Also using depth-images, a number of approaches have been proposed that extract manually designed features and discriminative machine learning models to predict joint locations in depth images or 3D joint-angles directly~\cite{choi2015collaborative,keskin2013real,sun2015cascaded,tang2015opening}. More recently a number of deep-learning models have been proposed that take depth images as input and regress 2D joint locations in multiple images ~\cite{sridhar2013interactive,tompson2014real} which are then used for optimization-based hand pose estimation. Others deploy convolutional neural networks (CNNs) in end-to-end learning frameworks to regress 3D hand poses from depth images, either directly estimating 3D joint configurations~\cite{oberweger2017,sinha2016deephand}, or estimating  joint-angles instead of Cartesian coordinates~\cite{oberweger2015}. Exploiting the depth information more directly, it has also been proposed to convert depth images into 3D multi-views \cite{ge2016robust} or volumetric representations \cite{ge20173d} before feeding them to a 3D CNN. Aiming at more mobile usage scenarios, recent work has proposed hybrid methods for hand-pose estimation from body-worn cameras under heavy occlusion~\cite{mueller2017real}. While the main focus lies on RGB imagery, our work is also capable of predicting hand pose configurations from depth images due to the multi-modal latent space.

Wan et al.~\cite{wan2017} is the most related work in spirit to ours. Like our work, they employ deep generative models (a combination of VAEs and GANs) to learn a latent space representation that regularizes the posterior prediction. Our method differs significantly in that we propose a theoretically grounded derivation of a \crossmodal training scheme based on the variational autoencoder~\cite{kingma2013} framework that allows for joint training of a single \crossmodal latent space, whereas~\cite{wan2017} requires training of two separate latent spaces, learning of a mapping function linking them and final end-to-end refinement.
Furthermore, we experimentally show that our approach reaches parity with the state-of-the-art in depth based hand pose estimation \emph{and} outperforms existing methods in the RGB case, whereas~\cite{wan2017} report only depth based experiments. In \cite{bouchacourt2016disco}, VAE is  also deployed for depth based hand pose estimation. However, their focus is minimising the dissimilarity coefficient between the true distribution and the estimated distribution.

\begin{figure*}
\begin{center}
\includegraphics[width=1.0\linewidth,clip,trim={0mm 2mm 0mm 2mm}]{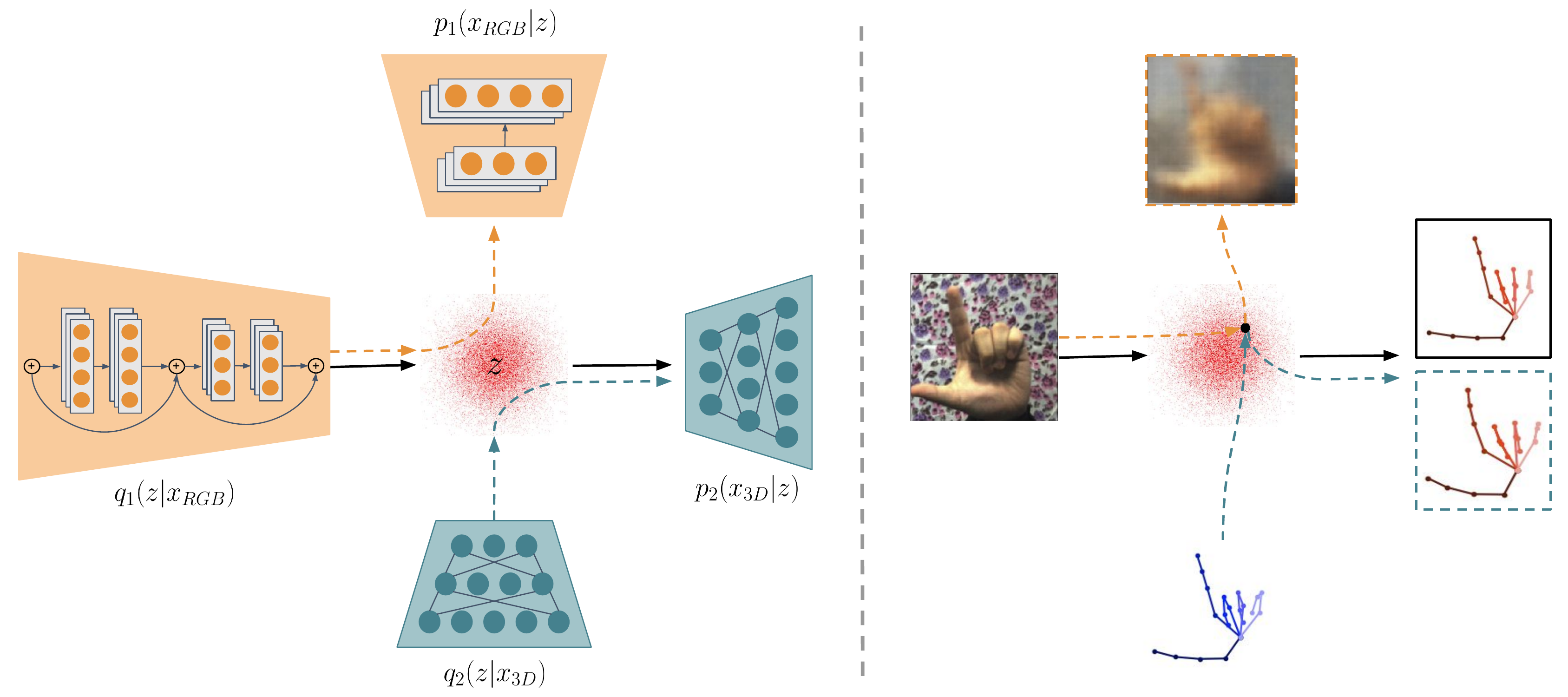}
\end{center}
\vspace*{-6mm}
   \caption{\textbf{Schematic overview of our architecture.} Left: a cross-modal latent space $z$ is learned by training pairs of encoder and decoder $q,p$ networks across multiple modalities (e.g., RGB images to 3D hand poses). Auxilliary encoder-decoder pairs help in regularizing the latent space. Right: The approach allows to embed input samples of one set of modalities (here: RGB, 3D) and to produce \emph{consistent} and plausible posterior estimates in several \emph{different} modalities (RGB, 2D and 3D).}
\label{fig:pipeline}
\end{figure*}

To the best of our knowledge there is currently only one approach for learning-based hand pose estimation from RGB images alone~\cite{zimmermann2017}. Demonstrating the feasibility of the task, this work splits 3D hand pose estimation into an image segmentation, 2D joint detection and 2D-3D lifting task. Our approach allows for training of the latent space using either input modality (in this case 2D key points or RGB images) and direct 3D hand pose estimation via decoding the corresponding sample from the latent space. We experimentally show that our methods outperforms~\cite{zimmermann2017} both in the 2D-3D lifting setting and the end-to-end hand pose estimation setting, even when using fewer invariances than the original method. Finally, we demonstrate that the same approach can be directly employed to depth images without any modifications to the architecture.

Our work builds on literature in deep generative modeling. Generative Adversarial Nets (GAN) \cite{goodfellow2014} learn an underlying distribution of the data via an adversarial learning process. The Variational Autoencoder (VAE) \cite{kingma2013} learns it via optimizing the log-likelihood of the data under a latent space manifold. However unlike GANs, they provide a framework to embed data into this manifold which has been shown to be useful for diverse applications such as multi-modal hashing \cite{erin2017}. Aytar et al. \cite{aytar2017} use several CNNs to co-embed data from different data modalities for scene classification and Ngiam et al. \cite{ngiam2011} reconstruct audio and video across modalities via a shared latent space. Our work also aims to create a \crossmodal latent space and we provide a derivation of the \crossmodal training objective function that naturally admits learning with different data sources all representing physically plausible hand pose configurations.

\section{Method}
The complex and dexterous articulation of the human hand is difficult to model directly with geometric or physical constraints \cite{oikonomidis2011efficient,taylor2016efficient,tzionas2016capturing}. However, there is broad agreement in the literature that a large amount of the degrees-of-freedom are not independently controllable and that hand motion, in natural movement, lives in a low-dimensional subspace~\cite{santello1998postural,todorov2004analysis}. Furthermore, it has been shown that dimensionality reduction techniques can provide data-driven priors in RGB-D based hand pose estimation~\cite{schroder2014real,tagliasacchi_sgp15}. However, in order to utilize such a low-dimensional sub-space directly for posterior estimation in 3D hand-pose estimation it needs to be
\begin{inparaenum}[i)]
    \item smooth,
    \item continuous and
    \item consistent.
\end{inparaenum}
Due to the inherent difficulties of capturing hand poses, most data sets do not cover the full motion space and hence the desired manifold is not directly attainable via simple dimensionality reduction techniques such as PCA.

We deploy the VAE framework that admits \crossmodal training of such a hand pose latent space by using various sources of data representation, even if stemming from different data sets both in terms of input and output. Our \crossmodal training scheme, illustrated in \figref{fig:pipeline}, learns to embed hand pose data from different modalities and to reconstruct them either in the same or in a different modality.

More precisely, a set of encoders $q$ take data samples $x$ in the form of either 2D keypoints, RGB or depth images and project them into a low-dimensional latent space $z$, representing physically plausible poses. A set of decoders $p$ reconstruct the hand configuration in either modality. The focus of our work is on 3D hand pose estimation and therefore on estimating the 3D joint posterior. The proposed approach is fully generative and experimentally we show that it is capable of generating consistent hand configurations across modalities. During training, each input modality alternatively contributes to the construction of the shared latent space. The manifold is continuous and smooth which we show by generating cross-modal samples such as novel pairs of 3D poses and images of natural hands\footnote{Generated images are legible but blurry. Creating high quality natural images is a research topic in itself.}.

\subsection{Variational Autoencoder}
Our \crossmodal training objective can be derived from the VAE framework \cite{kingma2013}, a popular class of generative models, typically used to synthesize data. A latent representation is attained via optimizing the so-called variational lower bound on the log-likelihood of the data:
\begin{equation}\label{eq:kld-vanilla}
\log p(x) \geq E_{z \sim q(z|x)}[\log p(x|z)] - D_{KL}(q(z|x)||p(z))
\end{equation}
Here $D_{KL}(\cdot)$ is the Kullback-Leibler divergence, and the conditional probability distributions $q(z|x)$, $p(x|z)$ are the encoder and decoders, parametrized by neural networks. The distribution $p(z)$ is the prior on the latent space,  modeled as $\mathcal{N}(z|0,I)$. The encoder returns the mean $\mu$ and variance $\sigma^2$ of a normal distribution, such that $z \sim \mathcal{N}(\mu, \sigma^2)$.

In this original form VAEs only take a single data distribution into account. To admit \crossmodal training, at least two data modalities need to be considered.

\subsection{Cross-modal Hand Pose Latent Space}
Our goal is to guide the \crossmodal VAE into learning a lower-dimensional latent space of hand poses with the above mentioned desired properties and the ability to project any modality into $z$ and to generate posterior estimates in any modality. For this purpose we re-derive a new objective function for training which leverages multiple modalities. We then detail our training algorithm based on this objective function.

For brevity we use a concrete example in which a data sample $x_i$ (e.g., an RGB image) is embedded into the latent space to obtain the embedding vector $z$, from which a corresponding data sample $x_t$ is reconstructed (e.g., a 3D joint configuration). To achieve this, we maximize the log-probability of our desired output modality $x_t$ under our model $\log p_{\theta}(x_t)$, where $\theta$ are the model parameters. We will omit the model parameters to reduce clutter.

Similar to the original derivation \cite{kingma2013}, we start with the quantity $\log p(x_t)$ that we want to maximize:
\begin{equation}
\label{eq:vae-eq-model}
\log p(x_t) = \int_zq(z|x_i)\log p(x_t)dz  ,
\end{equation}
exploiting the fact that $\int_zq(z|x_i)dz = 1$ and expanding $p(x_t)$ gives:
\begin{equation}
\label{eq:vae-eq-extended}
\int_zq(z|x_i)\log \frac{p(x_t)p(z|x_t)q(z|x_i)}{p(z|x_t)q(z|x_i)}dz.
\end{equation}
Remembering that $D_{KL}(p(x)||q(x))= \int_x p(x) \log \frac{p(x)}{q(x)}$ and splitting the integral of Eq~\eqref{eq:vae-eq-extended} we arrive at:
\begin{equation}
\begin{split}
& \int_zq(z|x_i)\log \frac{q(z|x_i)}{p(z|x_t)}dz + \int_zq(z|x_i)\log \frac{p(x_t)p(z|x_t)}{q(z|x_i)}dz\\
& = D_{KL}(q(z|x_i)||p(z|x_t)) + \int_zq(z|x_i)\log (\frac{p(x_t|z)p(z)}{q(z|x_i)})dz. 
\end{split}
\label{eq:vae-eq-kld}
\end{equation}
Here $p(z|x_t)$ corresponds to the desired but inaccessible posterior, which we approximate with $q(z|x_i)$.

Since $p(x_t)p(z|x_t) = p(x_t|z)p(z)$ and because $D_{KL}(p(x)||q(x)) \geq 0$ for any distribution $p,q$, we attain the final lower bound:
\begin{equation}
\begin{split}
& D_{KL}(q(z|x_i)||p(z|x_t)) + \int_zq(z|x_i)\log (\frac{p(x_t|z)p(z)}{q(z|x_i)})dz\\
& \geq \int_zq(z|x_i)\log p(x_t|z)dz - \int_zq(z|x_i)\log \frac{q(z|x_i)}{p(z)}dz\\
& = \mathbb{E}_{z \sim q(z|x_i)}[\log p(x_t|z)] - D_{KL}(q(z|x_i)||p(z)) .\\
\end{split}
\end{equation}
Note that we changed signs via the identity $-\log(x) = \log(\frac{1}{x})$. Here $q(z|x_i)$ is our encoder, embedding $x_i$ into the latent space and $p(x_t|z)$ is the decoder, which transforms the latent sample $z$ into the desired representation $x_t$.

The derivation shows that input samples $x_i$ and target samples $x_t$ can be decoupled via a joint embedding space $z$ where $i$ and $t$ can represent any modality.
For example, to maximize $\log p(x_\text{3D})$ when given $x_\text{RGB}$, we can train with $q(z|x_{\text{RGB}})$ as our encoder and $p(x_{\text{3D}}|z)$ as the decoder.

Importantly the above derivation also allows to train additional encoder-decoder pairs such as ($q(z|x_{\text{RGB}})$,\,$p(x_{\text{RGB}}|z)$), at the same time, for the same $z$.
This \crossmodal training regime results in a single latent space that allows us to embed and reconstruct multiple data modalities, or even train in a unsupervised fashion.

In the context of hand pose estimation, $p(z)$ represents a hand pose manifold which can be better defined with additional input modalities such as $x_\text{RGB}$, $x_\text{2D}$, $x_\text{3D}$, and even $x_\text{Depth}$ used in combination.

\subsection{Network Architecture}
In practice, the encoder $q_k$ for data modality $k$ returns the mean $\mu$ and variance $\sigma^2$ of a normal distribution for a given sample, from which the embedding $z$ is sampled, i.e $z \sim \mathcal{N}(\mu, \sigma^2)$. However, the decoder $p_l$ directly reconstructs the latent sample $z$ to the desired data modality $l$.

\figref{fig:pipeline}, illustrates our proposed architecture for the case of RGB based handpose estimation. In this setting we use two encoders for RGB images and 3D keypoints respectively.
Furthermore, the architecture contains two decoders for RGB images and 3D joint configurations.

\subsection{Training Procedure}
Our \crossmodal objective function (Eq~\ref{eq:vae-eq-extended}) follows the training procedure given as pseudo-code in Alg.\ref{alg:training}. The procedure takes a set of modalities $P_{VAE}$ with corresponding encoders and decoders $q_i,p_j$, where $i,j$ signify the respective modality, and trains all such pairs iteratively for $\mathcal{E}$ epochs. Note that the embedding space $z$ is always the same and hence we attain a joint \crossmodal latent space from this procedure (cf. \figref{fig:teaser}).

\begin{algorithm}
\caption{Cross-modal Variational Autoencoders}
\label{alg:training}
\begin{algorithmic}
    \STATE $P_{VAE} \leftarrow \{(q_{k_1},p_{l_1}), (q_{k_2},p_{l_2}), ...\}$ Encoder/Decoder pairs, where $q_{k_1}$ encodes data from modality $k_1$ and $p_{l_1}$ reconstructs latent samples to data of modality $l_1$.
    \STATE $\mathcal{E}$ Number of epochs
    \STATE $e \leftarrow 0$
    \FOR{$e < \mathcal{E}$}
      \FOR{$(q_k,p_l) \in P_{VAE}$}
          \STATE $x_k, x_l \leftarrow X_k, X_l$ Sample data pair of modality $k,l$
          \STATE $\mu, \sigma \leftarrow q_k(x_k)$
          \STATE $z \sim \mathcal{N}(\mu, \sigma)$
          \STATE $\hat{x}_l \leftarrow p_l(z)$
          \STATE $\mathcal{L}_{MSE} \leftarrow ||x_l - \hat{x}_l||_2$
          \STATE $\mathcal{L}_{KL} \leftarrow -0.5 * (1 +\log(\sigma^2) - \mu^2 - \sigma^2)$
          \STATE $\theta_{q_k} \leftarrow \theta_{q_k} - \nabla_{\theta_{q_k}} (\mathcal{L}_{MSE} + \mathcal{L}_{KL})$
          \STATE $\theta_{p_l} \leftarrow \theta_{p_l} - \nabla_{\theta_{p_l}} (\mathcal{L}_{MSE} + \mathcal{L}_{KL})$
      \ENDFOR
      \STATE $e \leftarrow e + 1$
    \ENDFOR
\end{algorithmic}
\end{algorithm}


\section{Experiments}
\label{sec:experiments}

To evaluate the performance of the \crossmodal VAE we systematically evaluate the utility of the proposed training algorithm and the resulting \crossmodal latent space. This is done via estimation of 3D hand joint positions from three entirely different input modalities: 1) 2D joint locations; 2) RGB image; 3) depth images.
In our experiments we explored combinations of different modalities during training. We always predict at least the 3D hand configuration but add further modalities.
More specifically we run experiments with the following four \textbf{variants}:
\begin{inparaenum}[a)]
\item \textbf{\emph{Var.~1}}: $(x_{i} \rightarrow x_{t})$
\item \textbf{\emph{Var.~2}}: $(x_{i} \rightarrow x_{t}, x_{t} \rightarrow x_{t})$
\item \textbf{\emph{Var.~3}}: $(x_{i} \rightarrow x_{t}, x_{i} \rightarrow x_{i})$
\item \textbf{\emph{Var.~4}}: $(x_{i} \rightarrow x_{t}, x_{i} \rightarrow x_{i}, x_{t} \rightarrow x_{t})$,
\end{inparaenum}
where $x_i$ always signifies the input modality and $i$ takes \textit{one} of the following values: [RGB, 2D, Depth]  and $t$ equals the output modality. In our experiments this is always $t=\text{3D}$ but can in general be any target modality. Including the $x_t \rightarrow x_i$  direction neither directly affects the RGB encoder, nor the 3D joint decoder and hence was dropped from our analysis.

\subsection{Implementation details}\label{sec:implementation-details}
We employ Resnet-18~\cite{he2016} for the encoding of RGB and depth images. Note that the model size of this encoder is much smaller compared to prior work that directly regresses 3D joint coordinates~\cite{oberweger2017}.
The decoders for RGB and depth consist of a series of (TransposedConv, BatchNorm2D and ReLU)-layers. For the case of 2D keypoint and 3D joint encoders and decoders, we use several (Linear, ReLU)-layers. In our experiments we did not observe much increase in accuracy from more complex decoder architectures. We train our architecture with the ADAM optimizer using a learning rate of $10^{-4}$. Exact architecture details and hyperparameters can be found in the supplementary materials.

\subsection{Datasets}
\label{sec:datasets}
We evaluate our method in the above settings based on several publicly available datasets.
For the input modality of \textbf{2D keypoints} and \textbf{RGB images} only few annotated datasets are available. We test on the datasets of the Stereo Hand Pose Tracking Benchmark \cite{zhang2016} (STB) and the Rendered Hand Pose Dataset (RHD) \cite{zimmermann2017}. STB contains $18$k images with resolution of $640\times 480$, which are split into a training set with $15$k samples and test set with $3$k samples. These images are annotated with 3D keypoint locations and the 2D keypoints are recovered via projecting them with the camera intrinsic matrix. The depicted hand poses contain little self-occlusion and variation in global orientation, lighting etc. and are relatively easy to recover.

RHD is a synthetic dataset with rendered hand images, which is composed of $42$k training images and $2.7$k evaluation images of size $320\times 320$. Similar to STB, both 2D and 3D keypoint locations are annotated. The dataset contains a much richer variety of viewpoints and poses. The 3D human model is set in front of randomly sampled images from Flickr to generate arbitrary backgrounds. This dataset is considerably more challenging due to variable viewpoints and difficult hand poses at different scales. Furthermore, despite being a synthetic dataset the images contain significant amount of noise and blur and are relatively low-res.

For the \textbf{depth} data, we evaluate on the ICVL \cite{tang2014}, NYU \cite{tompson2014real}, and MSRA \cite{sun2015cascaded} datasets.
For NYU, we train and test on viewpoint $1$ and all $36$ available joints, and evaluate on $14$ joints as done in \cite{oberweger2017,oberweger2015hands,wan2017} while for MSRA, we perform a leave-one-out cross-validation and evaluate the errors for the $9$ models trained as done in \cite{oberweger2017,sun2015cascaded,wan2017}.

\subsection{Evaluation metrics}
We provide three different metrics to evaluate the performance of our proposed model under various settings:
\begin{inparaenum}[i)]
\item The most common metric used in the 3D hand pose estimation literature is the \emph{mean 3D joint error} which measures the average euclidean distance between predicted joints and ground truth joints.
\item We also report \textbf{\emph{Percentage of Correct Keypoints} (PCK)} which returns the mean percentage of predicted joints below an euclidean distance of $d$ from the correct joint location.
\item The hardest metric, which reports the \textbf{\emph{Percentage of Correct Frames} (PCF)} where \emph{all} the predicted joints are within an euclidean distance of $d$ to its respective GT location. We report this only for depth since it is commonly reported in the literature.
\end{inparaenum}


\subsection{Comparison of variants}
We begin with comparing our variants with each other to determine which performs best and experiment on RHD and STB. On both datasets, we test the performance of our model on the task of regressing the 3D joints from RGB directly. Additionally, we predict the 3D joint locations from given 2D joint locations (dimensionality lifting) on RHD.

Table \ref{table:var_comparison} shows our results on the corresponding task and dataset. The errors are given in mean \textbf{end-point-error (EPE)} (median EPE is in the supplementary). Var.~3 outperforms the other variants on two tasks; lifting 2D joint locations to 3D on RHD and regressing 3D joint location directly from RGB on STB. On the other hand, Var.~1 is superior in the task of RGB$\rightarrow$3D on RHD. However we note that in general, the individual performance differences are minor. This is to be expected, as we conduct all our experiments within individual datasets. Hence even if multiple modalities are present, they capture the same poses and the same inherent information. This indicates that having a shared latent space for generative purposes does not harm the performance and in certain cases can even enhance it. This may be due to the regularizing effect of introducing multiple modalities.
\begin{table}
\centering
\begin{tabular}{|l|c|c|c|}
\hline
& \begin{tabular}{@{}c@{}}2D$\rightarrow$3D \\ RHD\end{tabular} & \begin{tabular}{@{}c@{}}RGB$\rightarrow$3D \\ RHD\end{tabular} & \begin{tabular}{@{}c@{}}RGB$\rightarrow$3D \\ STB\end{tabular} \\
\hline\hline
Var. 1 & 17.23 & \textbf{19.73} & 8.75 \\
Var. 2 & 17.82 & 19.99 & 8.61\\
Var. 3 & \textbf{17.14} & 20.04 & \textbf{8.56}\\
Var. 4 & 17.63 & 20.35 & 9.57\\
\hline
\end{tabular}
\caption{Variant comparison. Mean EPE given in mm. For explanation of variants, see Sec.~\ref{sec:experiments}.}
\label{table:var_comparison}
\end{table}


\subsection{Comparison to related work}
\label{sec:comparison_related_work}
In this section we perform a qualitative analysis of our performance in relation to prior work for both RGB and depth cases. For this, we pick the best variant of the respective task, as determined in the previous section. For the RGB datasets (RHD and STB), we compare against \cite{zimmermann2017}. To the best of our knowledge, it is the only prior work that addresses the same task as we do. In order to compare fairly, we conduct the same data preprocessing. Importantly, in \cite{zimmermann2017} additional information such as \textbf{handedness (H)} and \textbf{scale of the hand (S)} are provided at test time. Furthermore, the cropped hands are normalized to a roughly uniform size. Finally, they change the task from predicting the global 3D joint coordinates to estimating a palm-relative, \textbf{translation invariant (T)} set of joint coordinates by providing ground truth information of the palm center. In our case, the handedness is provided via a boolean flag directly into the model.

However, in order to assess the influence of our learned hand model we incrementally reduce the reliance on invariances which require access to ground-truth information. These results are shown alongside our main algorithm.

\textbf{2D to 3D.}
As a baseline experiment we compare our method to that of \cite{zimmermann2017} in the task of lifting 2D keypoints into a 3D hand pose configuration on the RHD dataset. Recently \cite{martinez2017} report that given a good 2D keypoint detector, lifting to 3D can yield surprisingly good results, even with simple methods in the case of 3D human pose estimation. Hand pose estimation is considerably more challenging task due to the more complex motion and flexibility of the human hand. Furthermore, \cite{zimmermann2017} provide a separate evaluation of their lifting component which serves as our baseline.

The first column of Table~\ref{table:rgb_prior_work_comparison} summarizes the mean squared end-point errors (EPE) for the RHD dataset. In general, our proposed model outperforms \cite{zimmermann2017} by a relatively large margin. The bottom rows of Table~\ref{table:rgb_prior_work_comparison} show results of ours \emph{without} the handedness invariance (H) and the scale invariance (S), we still surpass the accuracy of \cite{zimmermann2017}.
This suggests that our model indeed encodes physically plausible hand poses and that reconstructing the posterior from the embedding aids the hand pose estimation task.

\textbf{RGB to 3D.}
Here, we evaluate our method on the task of directly predicting 3D hand pose from RGB images, without intermediate 2D keypoint extraction. We run our model and \cite{zimmermann2017} on cropped RGB images for fair comparison.

Zimmermann et al. \cite{zimmermann2017}, in which 2D keypoints are first predicted and then lifted into 3D serves as our baseline.  We evaluate the proposed model on the STB \cite{zhang2016} and RHD \cite{zimmermann2017} datasets. Fig.~\ref{fig:res_rgb_real} and \ref{fig:res_rgb_synth} show several samples of our prediction on STB and RHD respectively. Even though some images in RHD contain heavily occluded fingers, our method retrieves biomechanically plausible predictions.

\begin{figure}[t]
\begin{subfigure}[t]{0.48\columnwidth}
\centering
\includegraphics[width=\linewidth]{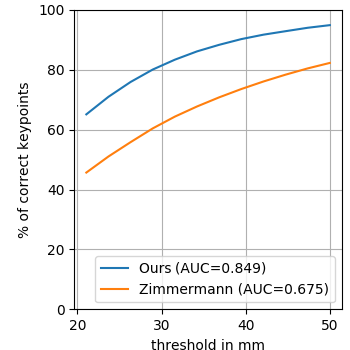}
\caption{RHD}
\label{fig:rgb23d_RHD_ex}
\end{subfigure}
\begin{subfigure}[t]{0.48\columnwidth}
\centering
   \includegraphics[width=\linewidth]{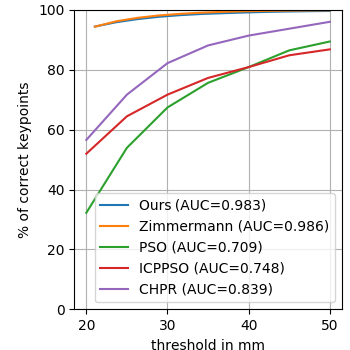}
\caption{STB}
\label{fig:rgb23d_STB_ex}
\end{subfigure}
\vspace*{-2mm}
\caption{PCK curve of our best model on RHD and STB for RGB to 3D. }
\end{figure}

The middle column of Table~\ref{table:rgb_prior_work_comparison} summarizes the results for the harder RHD dataset. Our approachs accuracy exceeds that of \cite{zimmermann2017} by a large margin. Removing available invariances again slightly decreases performance but our models still remains superior to \cite{zimmermann2017}. Looking at the PCK curve comparison in Fig. \ref{fig:rgb23d_RHD_ex}, we see that our model outperforms \cite{zimmermann2017} for all thresholds.

The rightmost column of Table~\ref{table:rgb_prior_work_comparison} shows the performance on the STB dataset. The margin of improvement of our approach is considerably smaller. We argue that the performance on the dataset is saturated as it is much easier (see discussion in Sec. \ref{sec:datasets}). Fig. \ref{fig:rgb23d_STB_ex} shows the PCK curves on STB, with the other baselines that operate on noisy stereo depth maps and \emph{not} RGB (directly taken from \cite{zimmermann2017}).

\begin{table}
\centering
\begin{tabular}{|l|c|c|c|}
\hline
& \begin{tabular}{@{}c@{}}2D$\rightarrow$3D \\ RHD\end{tabular} & \begin{tabular}{@{}c@{}}RGB$\rightarrow$3D \\ RHD\end{tabular} & \begin{tabular}{@{}c@{}}RGB$\rightarrow$3D \\ STB\end{tabular} \\
\hline\hline
\cite{zimmermann2017} (T+S+H) & 22.43 & 30.42 & 8.68\\ \hline
Ours (T+S+H) & \textbf{17.14} & \textbf{19.73} & \textbf{8.56}\\
Ours (T+S) & 18.90 & 20.20 & 10.16 \\
Ours (T+H) & 19.69 & 22.34 & 9.59 \\
Ours (T) & 21.15 & 22.53 & 9.49 \\
\hline
\end{tabular}
\vspace*{-3mm}
\caption{Related work comparison. Mean EPE given in mm. For explanation of legends, see Sec.~\ref{sec:comparison_related_work}}
\label{table:rgb_prior_work_comparison}
\end{table}

\begin{figure}
\centering
\begin{subfigure}{0.34\columnwidth}
  \includegraphics[width=\linewidth]{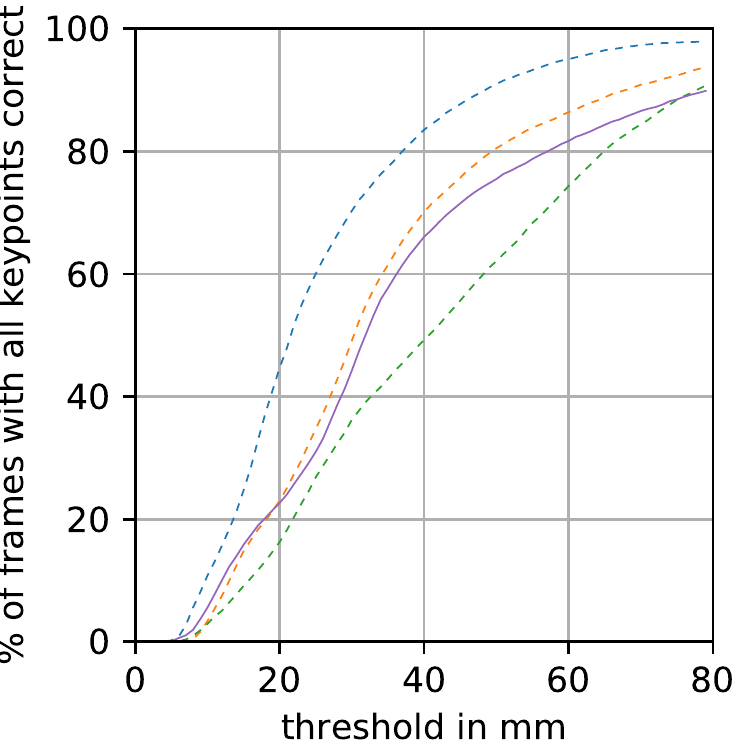}
  \caption{NYU}
\end{subfigure}
\hfill
\begin{subfigure}{0.32\columnwidth}
  \includegraphics[trim={5mm 0 0 0},clip,width=\linewidth]{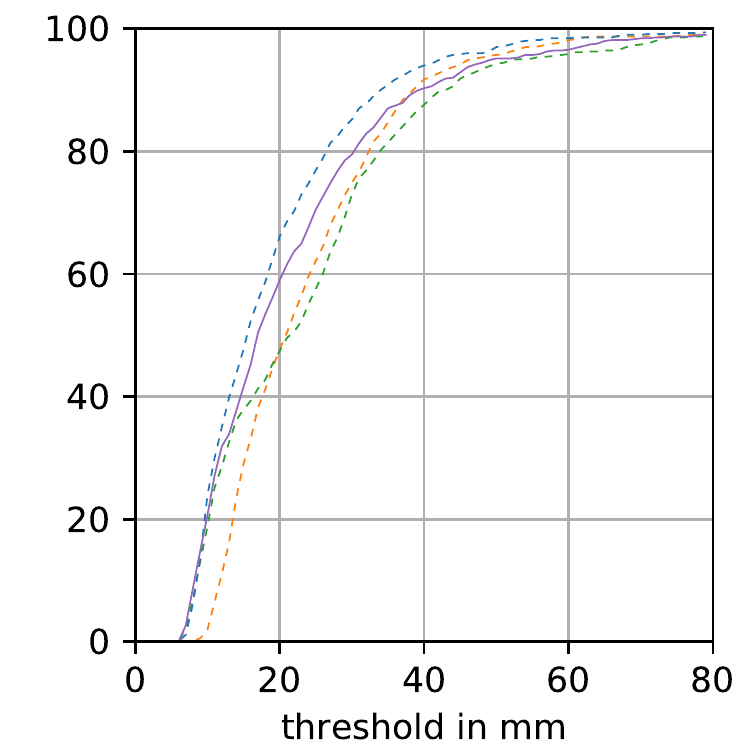}
  \caption{ICVL}
\end{subfigure}
\hfill
\begin{subfigure}{0.32\columnwidth}
  \includegraphics[trim={5mm 0 0 0},clip,width=\linewidth]{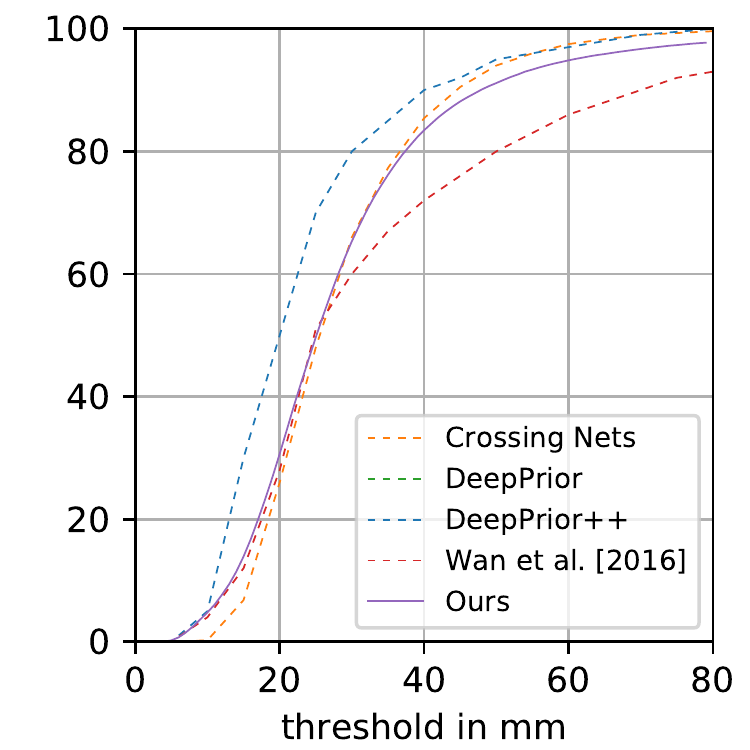}
  \caption{MSRA}
\end{subfigure}
\caption{PCF curves for 3D joint estimation from depth input. Our model performs comparably to recent works.}
\label{fig:depth_pcf}
\end{figure}

\textbf{Depth to 3D.}
Given the ready availability of RGB-D cameras, the task of 3D joint position estimation from depth has been explored in great detail and specialized architectures have been proposed. We evaluate our architecture, designed originally for the RGB case, on the ICVL \cite{tang2014}, NYU \cite{tompson2014real} and MSRA \cite{sun2015cascaded} datasets.
Despite the lower model capacity, our method performs comparably (see Fig.~\ref{fig:depth_pcf}) to recent works \cite{oberweger2017,oberweger2015hands,wan2017,wan2016} with just a modification to take $1$-channel images as input compared to our RGB case.

\subsection{Semi-supervised learning}
Due to the nature of cross-training, we can exploit complementary information from additional data. For example, if additional unlabeled images are available, our model can make use of these via cross-training. This is a common scenario, as unlabeled data is plentiful. If not available, acquiring this is by far simpler than recording training data.

\begin{figure}[t]
\centering
\includegraphics[width=0.8\linewidth]{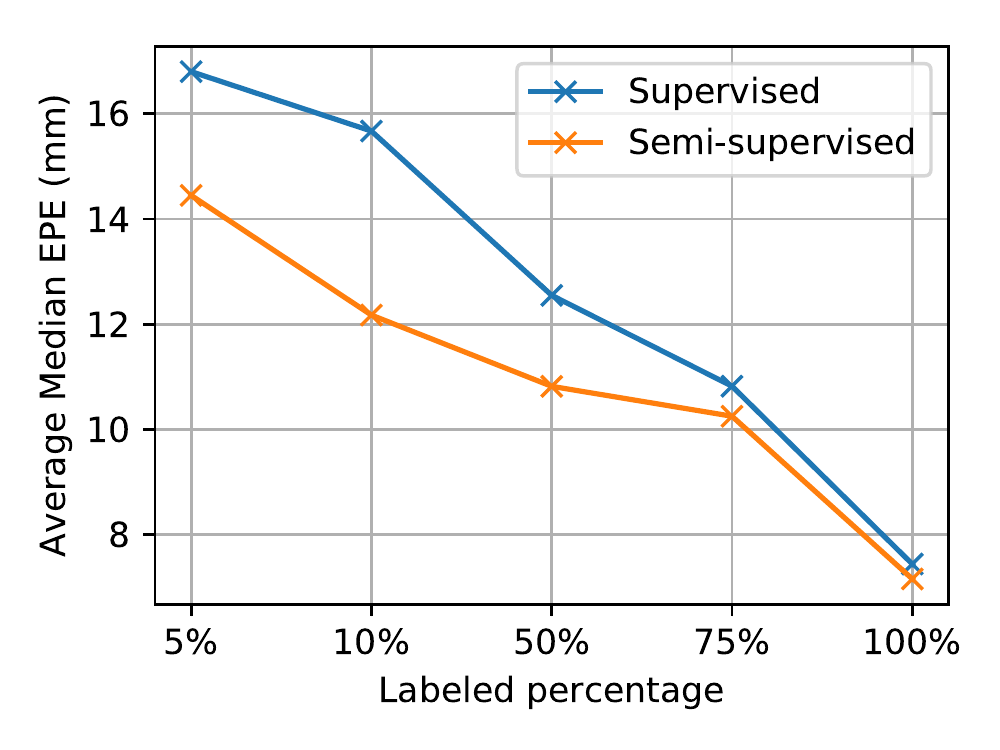}
\vspace*{-5mm}
\caption{Median EPE of our model trained supervised and semi-supervised as a function of percentage of labeled data.}
\label{fig:semsup_sup}
\end{figure}

To explore this semi-supervised setting, we perform an additional experiment on STB. We simulate a situation where we have labeled and unlabeled data by discarding different percentages of 3D joint data from our dataset. Fig. \ref{fig:semsup_sup}, compares the median EPE of Var.~1 (which can only be trained supervised) with Var.~3 (trained semi-supervised). We see that as more unlabeled data becomes available, Var.~3 can make use of this additional information and improve prediction accuracy up to 22$\%$.

\subsection{Generative capabilities}
\begin{figure*}[h!]
\begin{center}
\includegraphics[width=0.95\linewidth,clip,trim={0 52mm 0 8mm}]{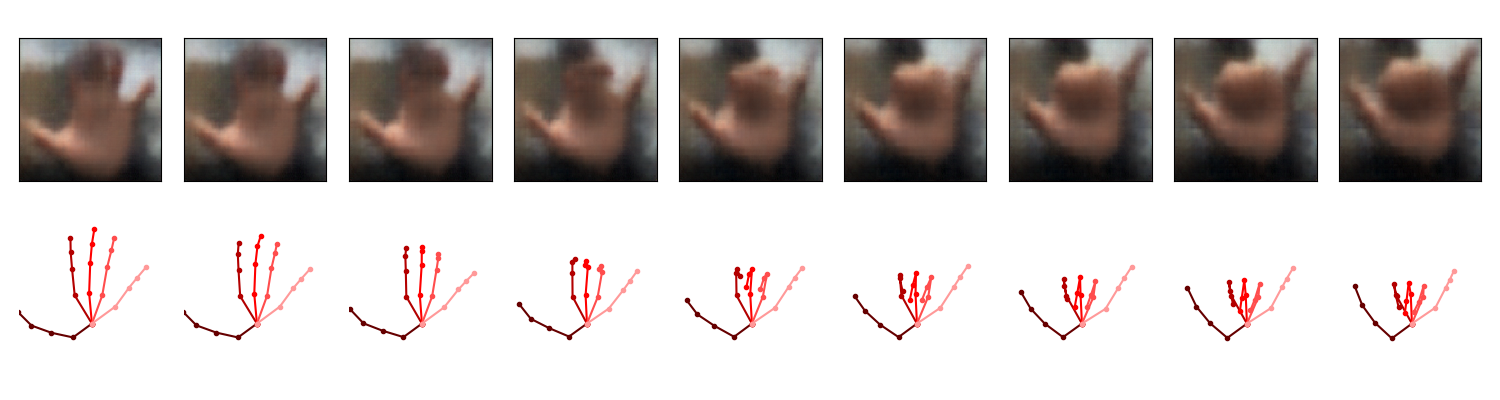}
\includegraphics[width=0.95\linewidth,clip,trim={0 13mm 0 58mm}]{./figures/latent_space_interpolation_wide_1.png}
\end{center}
\vspace*{-4mm}
\caption{\textbf{Latent space walk}. Example of reconstructing samples of the latent space into multiple modalities. The left-most and right-most figures are reconstruction from latent space samples of two real RGB images. The figures in-between are multi-modal reconstruction from interpolated latent space samples, hence are completely synthetic.}
\label{fig:latent_space_walk}
\end{figure*}

Our model is guided to learn a manifold of hand poses. In this section, we demonstrate the smoothness and consistency of it. To this end, we perform a walk on one dimension of the latent space by embedding two RGB images of separate hand poses into the latent space and obtain two corresponding samples $z_1$ and $z_2$. We then decode the latent space samples that reside on the interpolation line between them using our models for RGB and 3D joint decoding. Fig. \ref{fig:latent_space_walk} shows the resulting reconstructions, demonstrating consistency between both decoders. The fingers move in synchrony and the generated synthetic samples are both physically plausible and consistent across modalities. This demonstrates that the learned latent space is indeed smooth and represents a valid statistical model of hand poses.

The smoothness property of the unified latent space is attractive in several regards. Foremost because this potentially enables generation of \emph{labeled} data which in turn may be used to improve current models. Fully exploring this aspect is subject to further research.

\begin{figure*}[th!]
\newcommand{\stbvis}[3]{
  \includegraphics[trim={12mm 32mm 252mm 28mm},clip,width=0.10\linewidth]{#1}
  \includegraphics[trim={{\dimexpr 102mm-#2mm\relax} {\dimexpr 32mm-#3mm\relax} {\dimexpr 162mm+#2mm\relax} {\dimexpr 28mm+#3mm\relax}},clip,width=0.10\linewidth]{#1}
  \includegraphics[trim={{\dimexpr 224mm-#2mm\relax} {\dimexpr 32mm-#3mm\relax} {\dimexpr 40mm+#2mm\relax} {\dimexpr 28mm+#3mm\relax}},clip,width=0.10\linewidth]{#1}
}
\newcommand{\rhdvis}[3]{
  \includegraphics[trim={12mm 32mm 252mm 28mm},clip,width=0.10\linewidth]{#1}
  \includegraphics[trim={{\dimexpr 102mm-#2mm\relax} {\dimexpr 32mm-#3mm\relax} {\dimexpr 162mm+#2mm\relax} {\dimexpr 28mm+#3mm\relax}},clip,width=0.10\linewidth]{#1}
  \includegraphics[trim={{\dimexpr 224mm-#2mm\relax} {\dimexpr 32mm-#3mm\relax} {\dimexpr 40mm+#2mm\relax} {\dimexpr 28mm+#3mm\relax}},clip,width=0.10\linewidth]{#1}
}
\begin{subfigure}[t]{\linewidth}
\begin{center}
	\stbvis{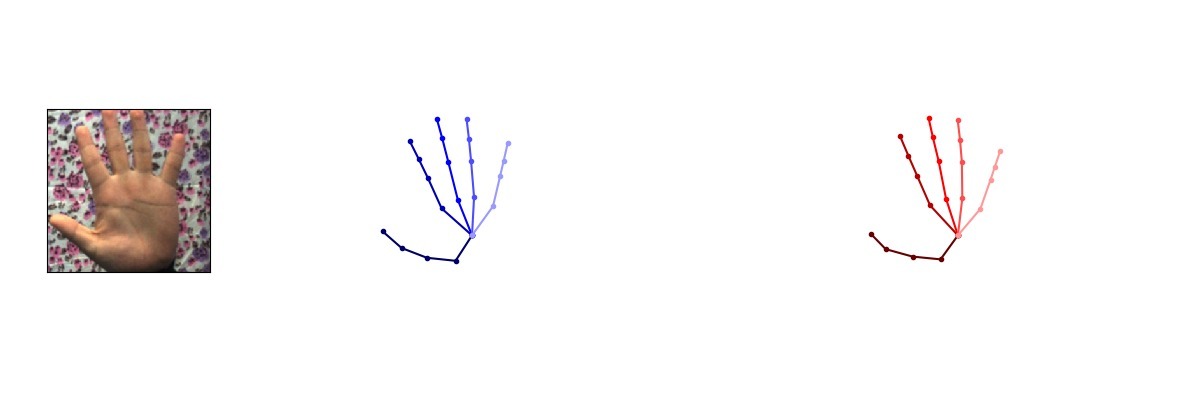}{6}{0}
	\stbvis{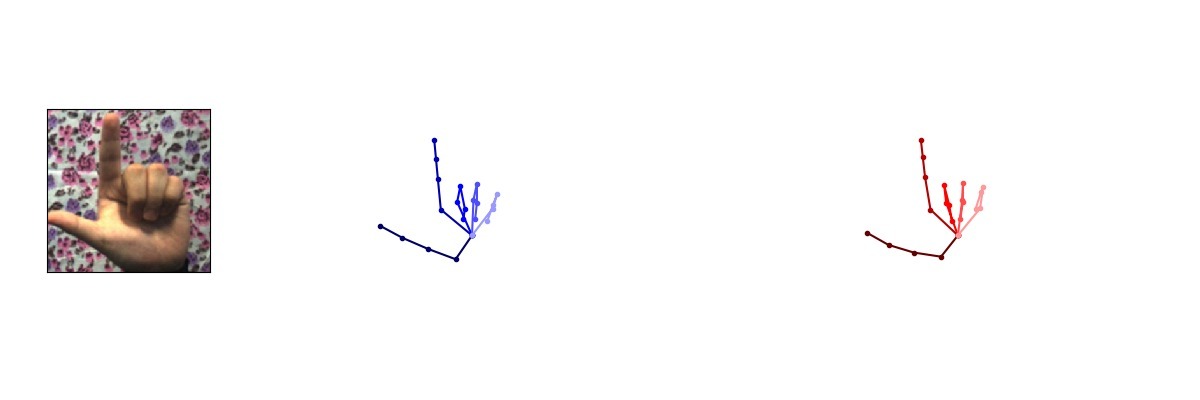}{8}{6}
	\stbvis{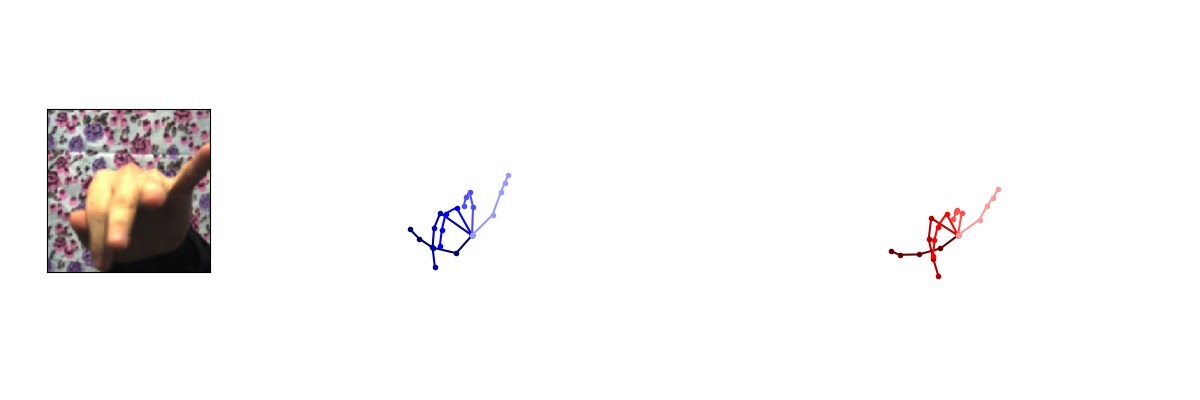}{6}{10}
	\stbvis{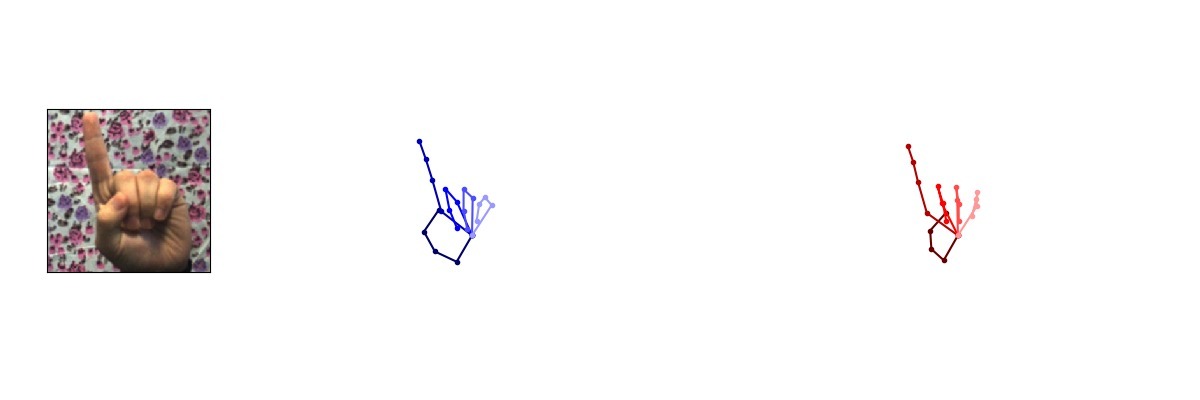}{6}{0}
	\stbvis{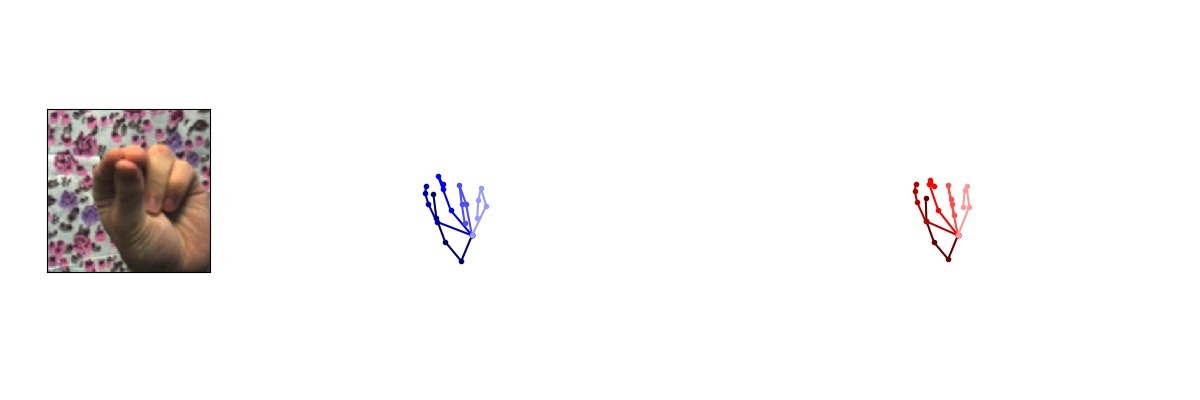}{6}{8}
	\stbvis{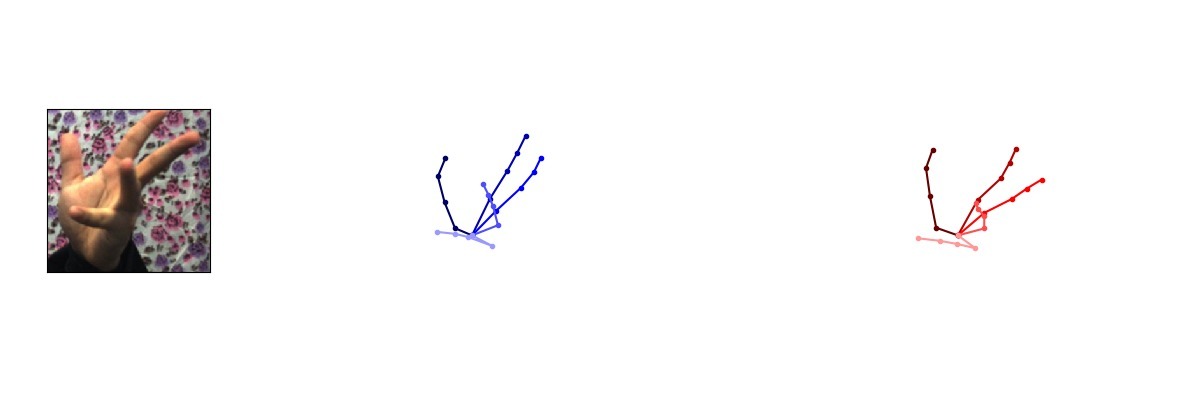}{3}{4}
\end{center}
\vspace*{-5mm}
\caption{\textbf{STB} (from RGB)}
\label{fig:res_rgb_real}
\end{subfigure}
\vspace{1mm}
\begin{subfigure}[t]{\linewidth}
\begin{center}
	\rhdvis{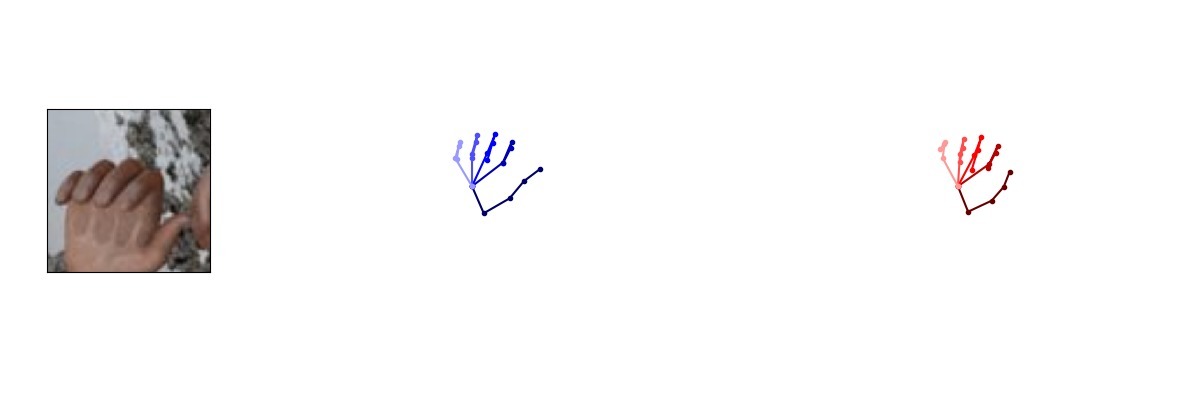}{-4}{-5}
	\rhdvis{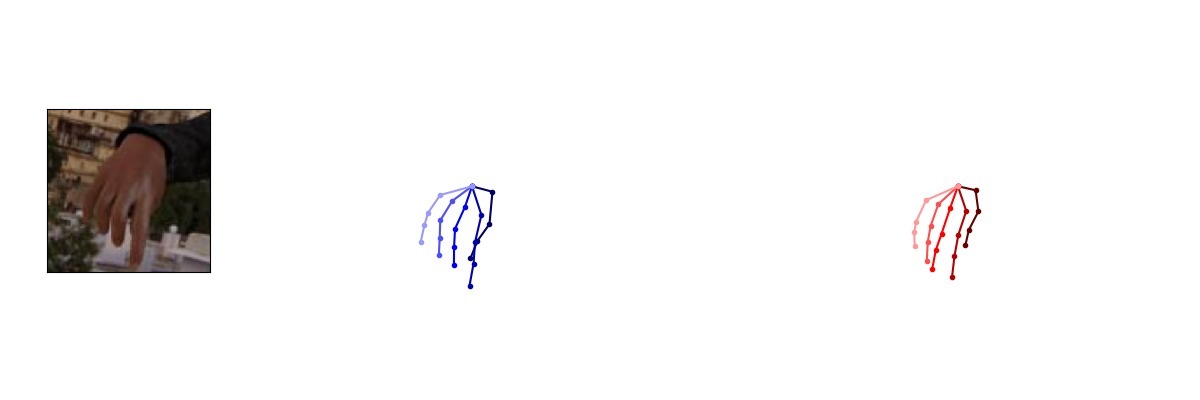}{4}{8}
	\rhdvis{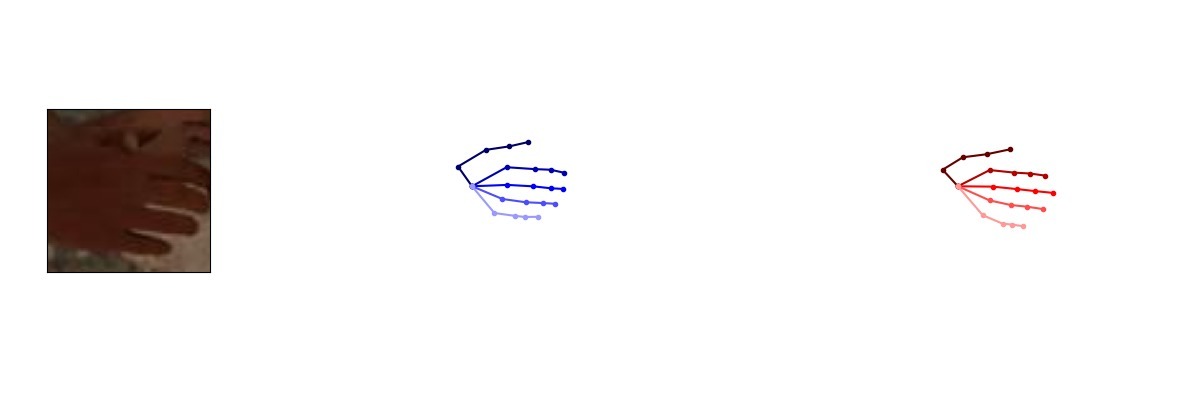}{-6}{-2}\\
	\rhdvis{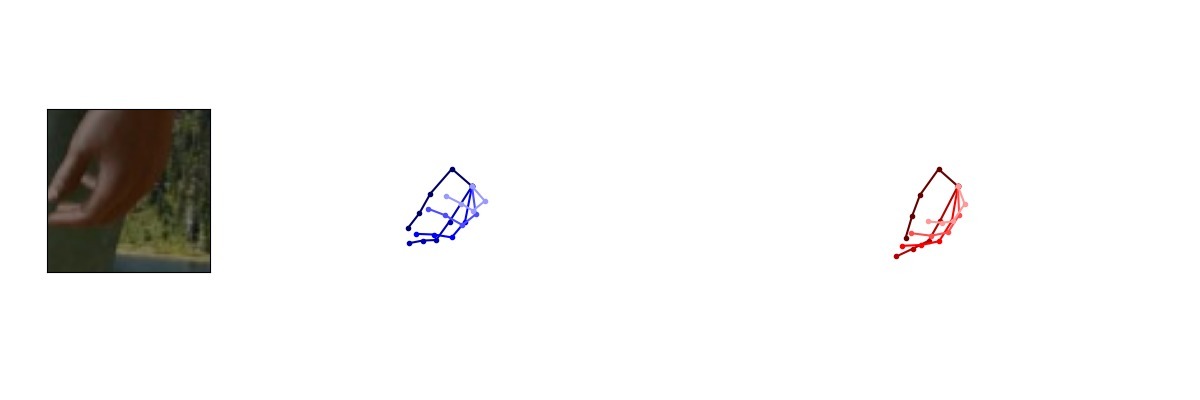}{7}{8}
	\rhdvis{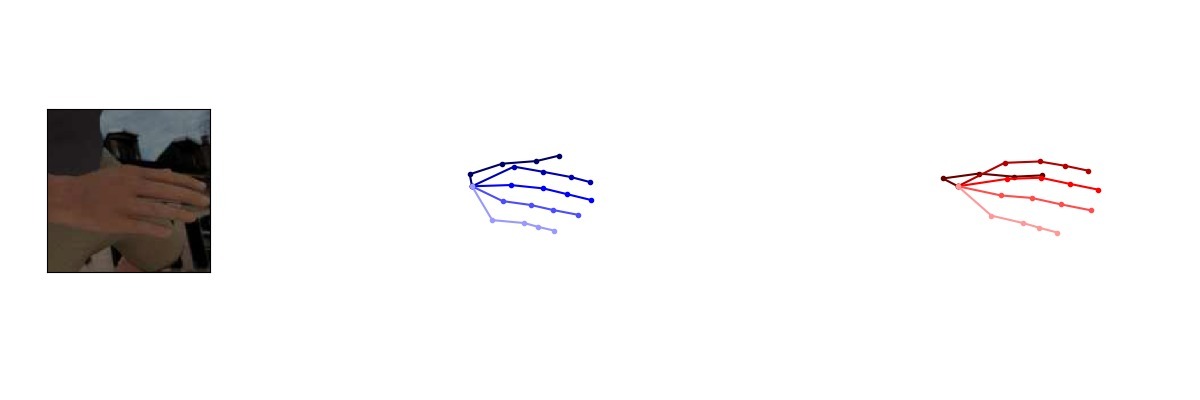}{-12}{0}
	\rhdvis{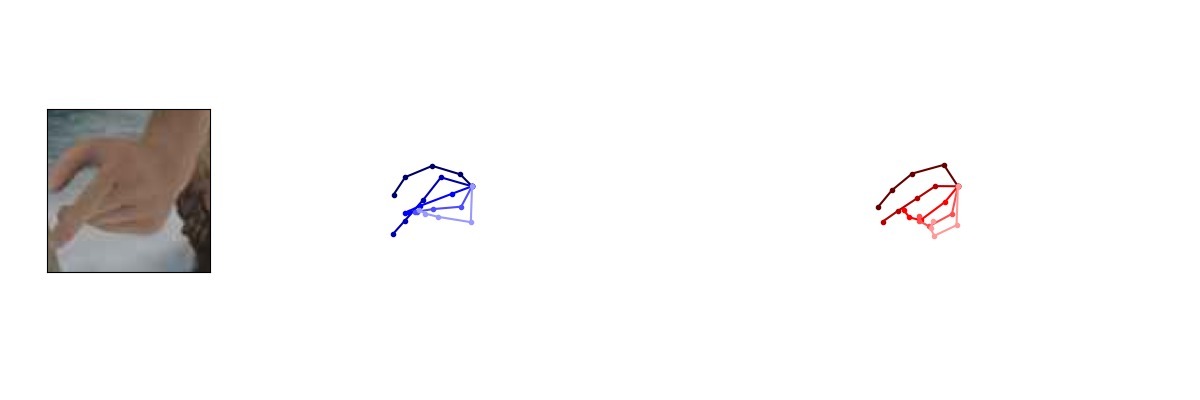}{12}{2}
\end{center}
\vspace*{-4mm}
\caption{\textbf{RHD} (from RGB)}
\label{fig:res_rgb_synth}
\end{subfigure}
\vspace{1mm}
\begin{subfigure}[t]{\linewidth}
\centering
\includegraphics[width=0.32\linewidth,clip,trim={0 18mm 0 8mm}]{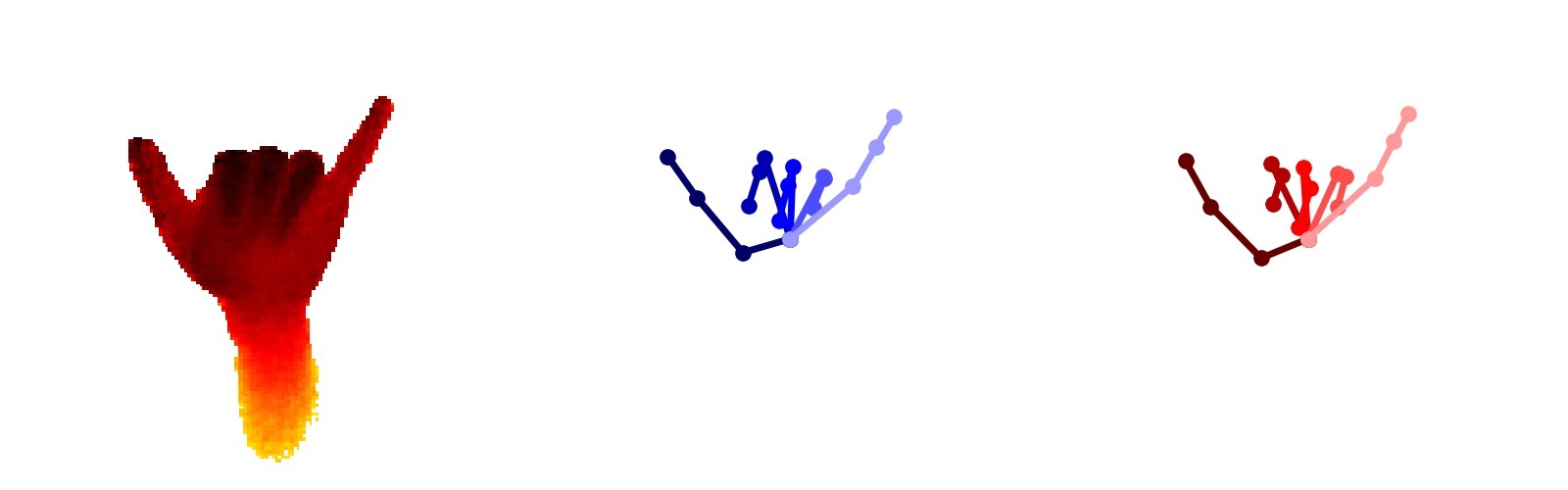}
\includegraphics[width=0.32\linewidth,clip,trim={0 18mm 0 8mm}]{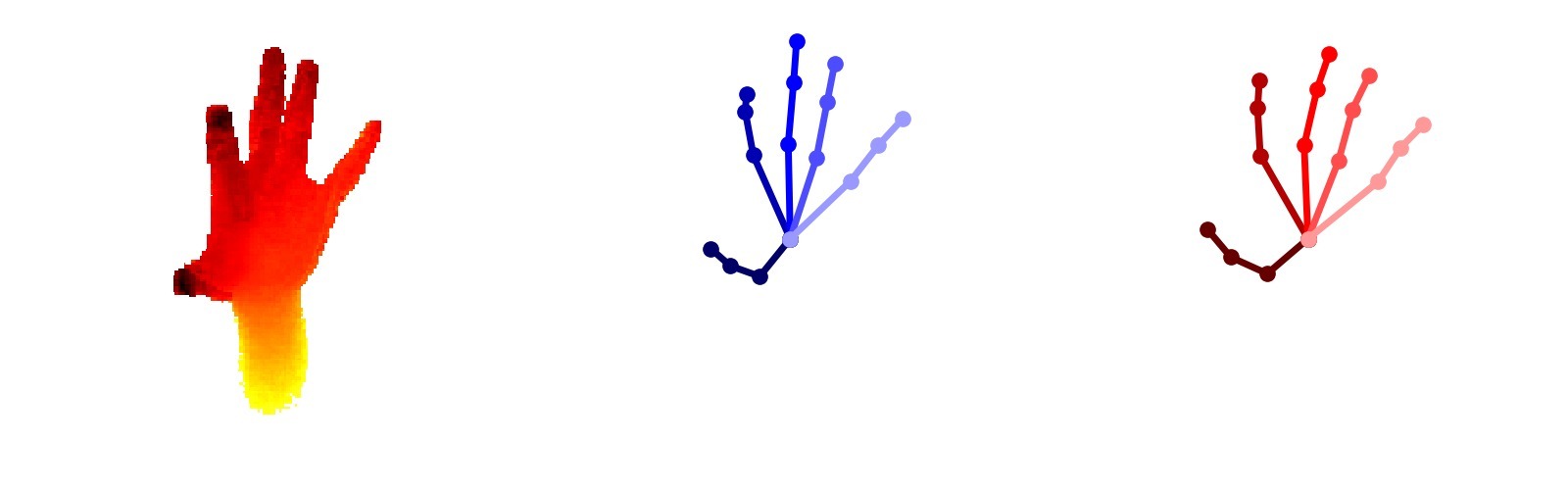}
\includegraphics[width=0.32\linewidth,clip,trim={0 18mm 0 8mm}]{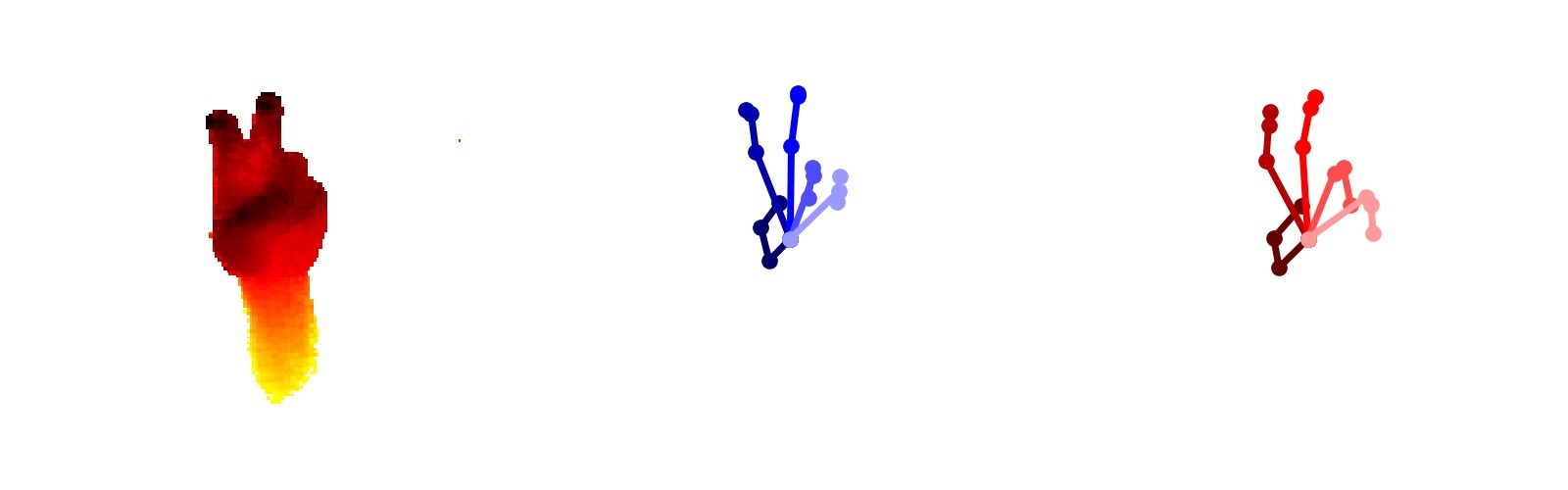}
\includegraphics[width=0.32\linewidth,clip,trim={0 18mm 0 8mm}]{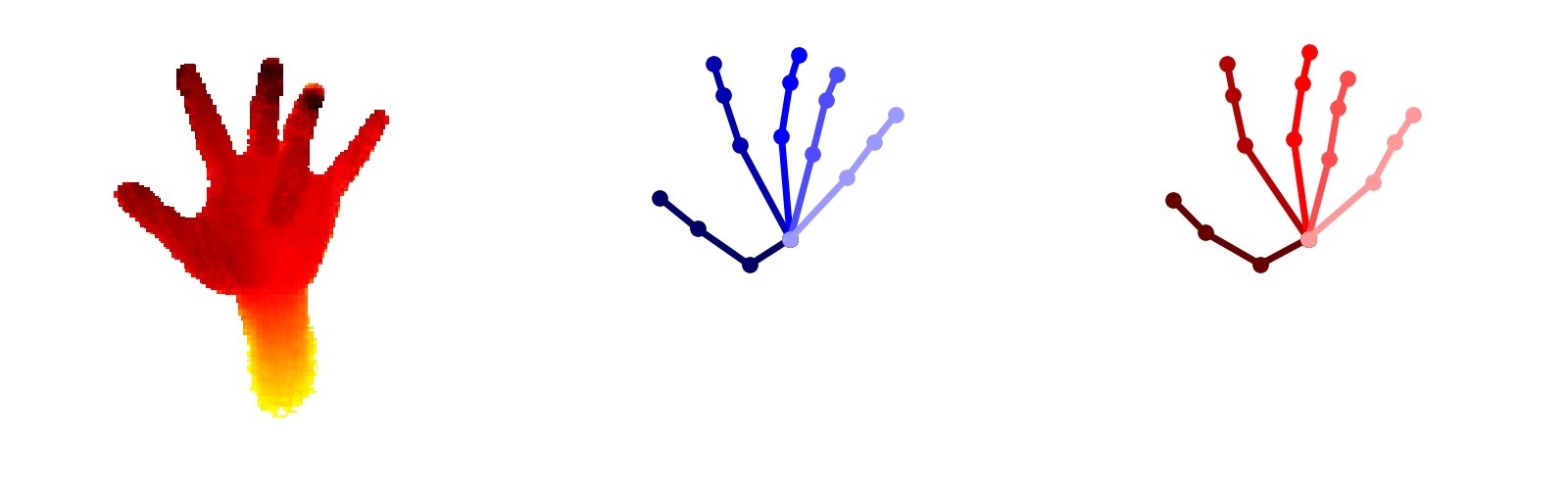}
\includegraphics[width=0.32\linewidth,clip,trim={0 18mm 0 8mm}]{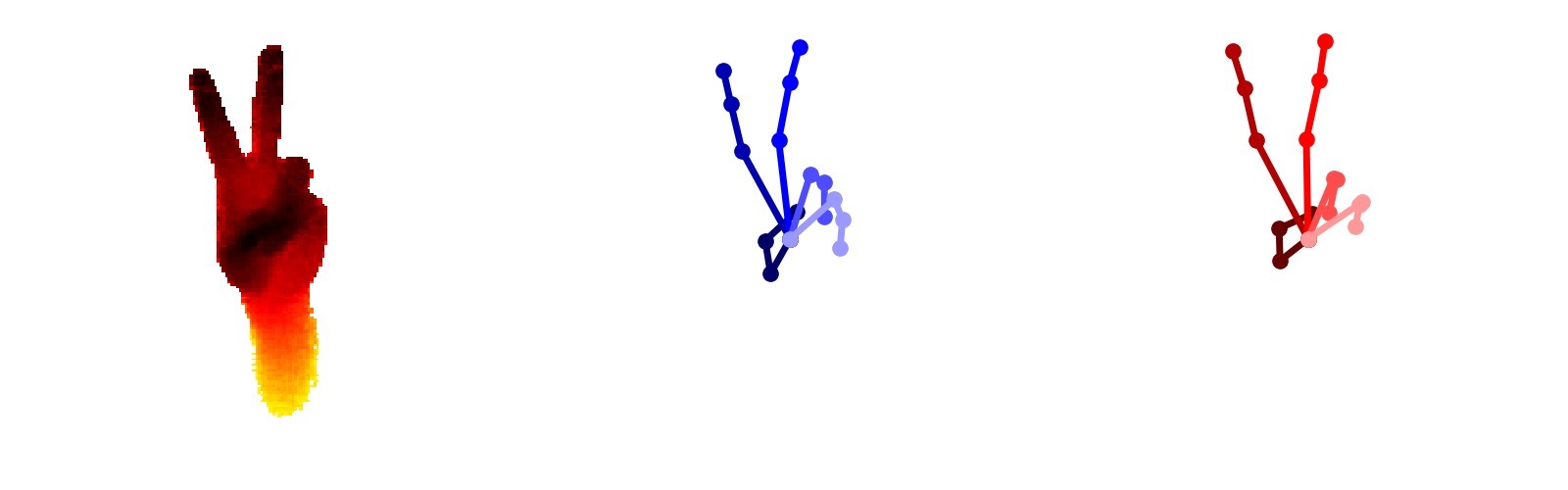}
\includegraphics[width=0.32\linewidth,clip,trim={0 18mm 0 8mm}]{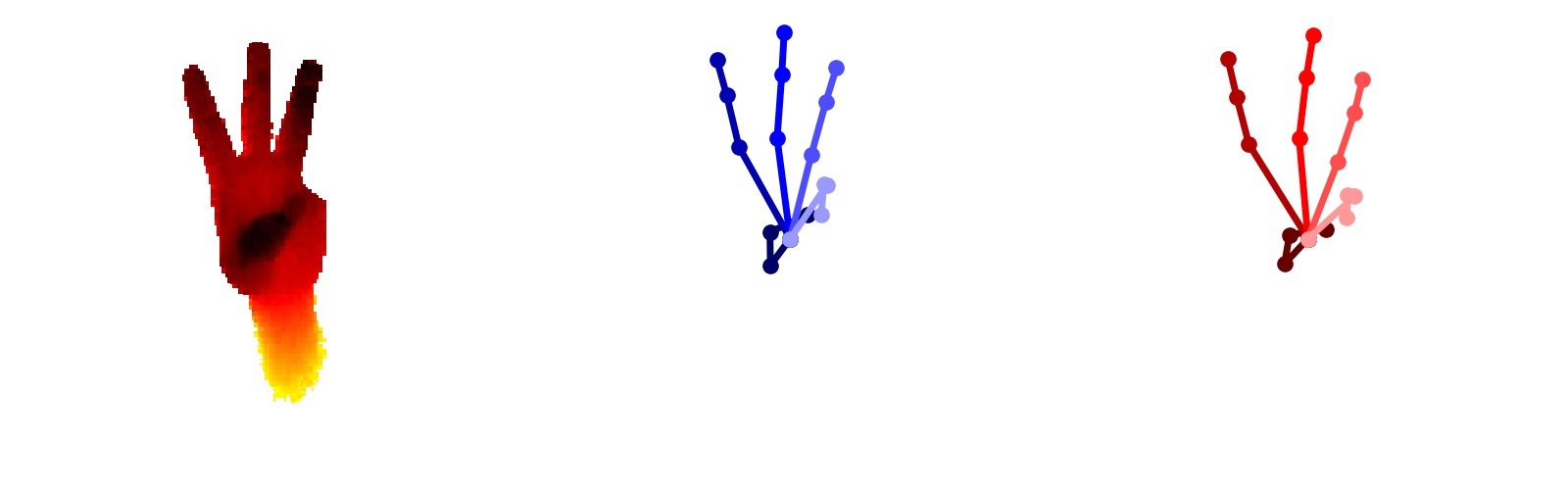}
\vspace*{-3mm}
\caption{\textbf{ICVL} (from Depth)}
\label{fig:res_depth}
\end{subfigure}
\vspace*{-4mm}
\caption{\textbf{3D joint predictions}. For each triplet, the left most column corresponds to the input image, the middle column is the ground truth 3D joint skeleton and the right column is our corresponding prediction.}
\end{figure*}


\section{Conclusion}
We have proposed a new approach to estimate 3D hand pose configurations from RGB and depth images. Our approach is based on a re-derivation of the variational lower bound that admits training of several independent pairs of encoders and decoders, shaping a joint cross-modal latent space representation. We have experimentally shown that the proposed approach outperforms the state-of-the art on publicly available RGB datasets and is at least comparable to highly specialized state-of-the-art methods on depth data. Finally, we have shown the generative nature of the approach which suggests that we indeed learn a usable and physically plausible statistical hand model, enabling direct estimation of the 3D joint posterior.

\section{Acknowledgements} This  work  was  supported  in  parts  by  the  ERC  grant OPTINT (StG-2016-717054)
\clearpage

{\small
\bibliographystyle{ieee}
\balance
\bibliography{egbib}
}
\newpage
\section{Supplementary}
This documents provides additional information regarding our main paper and discusses architecture, training and further implementation details. Furthermore, we provide additional experimental results in particular those that illustrate the benefit of the cross-modal latent space representation.
\subsection{Training details}
All code was implemented in PyTorch. For all models, we used the ADAM optimizer with its default parameters to train and set the learning rate of $10^{-4}$. The batch size was set to $64$.

\textbf{2D to 3D.}
For the 2D to 3D modality we use identical encoder and decoder architectures, consisting of a series of (Linear,ReLU)-layers. The exact architecture is summarized in table \ref{tbl:joint_encdec_arch}.

\begin{table}
\centering
\begin{tabular}{|c|c|}
\hline
\multicolumn{2}{|c|}{Encoder/Decoder}\\
\hline\hline
Linear($512$) & ReLU\\ \hline
Linear($512$) & ReLU\\ \hline
Linear($512$) & ReLU\\ \hline
Linear($512$) & ReLU\\ \hline
Linear($512$) & ReLU\\  \hline
\multicolumn{2}{|c|}{Linear($512$)}\\
\hline
\end{tabular}
\caption{Encoder and decoder architecture.}
\label{tbl:joint_encdec_arch}
\end{table}
\textbf{RGB to 3D.}
For the RGB to 3D modality, images were normalized to the range $[-0.5, 0.5]$ and we used data augmentation to increase the dataset size. More specifically, we randomly shifted the bounding box around the hand image, rotated the cropped images in the range $[-45^\circ,45^\circ]$ and applied random flips along the $y$-axis. The resulting image was then resized to $256\! \times \!256$. The joint data was augmented accordingly.\newline
Because the RHD and STB datasets have non-identical hand joint layouts (RHD gives the wrist-joint location, whereas STB gives the palm-joint location), we shifted the wrist joint of RHD into the palm via interpolating between the wrist and first middle-finger joint. We trained on both hands of the RHD dataset, whereas we used both views of the stereo camera of the STB dataset. This is the same procedure as in \cite{zimmermann2017}. The encoder and decoder architectures for RGB data are detailed in table \ref{tbl:rgb_encdec_arch}. We used the same encoder/decoder architecture for the 3D to 3D joint modality as for the 2D to 2D case (shown in table \ref{tbl:joint_encdec_arch}).

\textbf{Depth to 3D.} We used the same architecture and training regime as for the RGB case. The only difference was adjusting the number of input channels from $3$ to $1$.

\begin{table}
\centering
\begin{tabular}{|c|c|c|c|}
\hline
Encoder & \multicolumn{3}{c|}{Decoder}\\
\hline\hline
\multirow{7}{*}{\begin{turn}{-90}ResNet-18\end{turn}} & Linear($4096$) & BatchNorm & ReLU  \\ \cline{2-4}
& \multicolumn{3}{c|}{Reshape($256$, $4$, $4$)}  \\ \cline{2-4}
& ConvT($128$) & BatchNorm & ReLU \\ \cline{2-4}
&  ConvT($64$) & BatchNorm & ReLU \\ \cline{2-4}
&  ConvT($32$) & BatchNorm & ReLU \\ \cline{2-4}
&  ConvT($16$) & BatchNorm & ReLU  \\ \cline{2-4}
& ConvT($8$) & BatchNorm & ReLU  \\ \cline{2-4}
& \multicolumn{3}{c|}{ConvT($3$)} \\
\hline
\end{tabular}
\caption{Encoder and Decoder architecture for RGB data. ConvT corresponds to a layer performing transposed Convolution. The number indicated in the bracket is the number of output filters. Each ConvT layer uses a $4\times4$ kernel, stride of size $2$ and padding of size $1$.}
\label{tbl:rgb_encdec_arch}
\end{table}

\subsection{Qualitative Results}
In this section we provide additional qualitative results, all were produced with the architecture and training regime detailed in the main paper.

\begin{figure*}[h!]
\begin{subfigure}{.5\textwidth}
  \centering
  \includegraphics[width=1.05\linewidth]{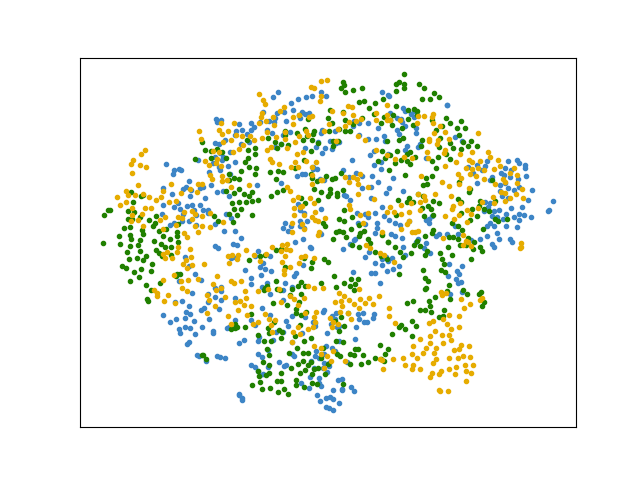}
  \caption{Cross-trained.}
  \label{fig:tsne_embedding_ct}
\end{subfigure}%
\hspace*{-5mm}
\begin{subfigure}{.5\textwidth}
  \centering
  \includegraphics[width=1.05\linewidth]{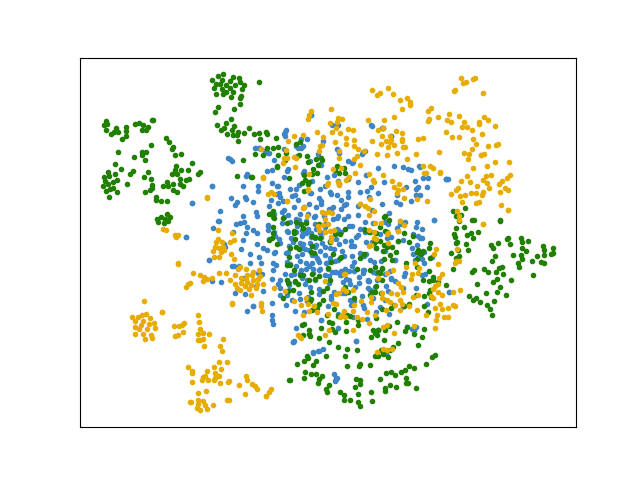}
  \caption{Not cross-trained.}
  \label{fig:tsne_embedding_noct}
\end{subfigure}
\caption{\textbf{t-SNE embedding of multi-modal latent space}. The two figures show the embedding of data samples from different modalities (blue: RGB images, green: 3D joints, yellow: 2D joints). In the left figure, our model was cross-trained, whereas in the right figure, each data modality was trained separately. This shows that in order to learn a multi-modal latent space, cross-training is vital.}
\label{fig:tsne_embedding}
\end{figure*}

\textbf{Latent space consistency.}
In Fig.~\ref{fig:tsne_embedding} we embed data samples from RHD and STB into the latent space and perform a t-SNE embedding. Each data modality is color coded (blue: RGB images, green: 3D joints, yellow: 2D joints). Here, Fig.~\ref{fig:tsne_embedding_ct} displays the embedding for our model when it is cross-trained. We see that each data modality is evenly distributed, forming a single, dense, approximately Gaussian cluster. Compared to Fig. \ref{fig:tsne_embedding_noct} which shows the embedding for the same model without cross-training, it is clear that each data modality lies on a separate manifold. This figure indicates that cross-training is vital for learning a multi-modal latent space.

\begin{figure*}[h!]
\begin{subfigure}[t]{\linewidth}
\begin{center}
\includegraphics[width=0.95\linewidth,clip,trim={0 52mm 0 8mm}]{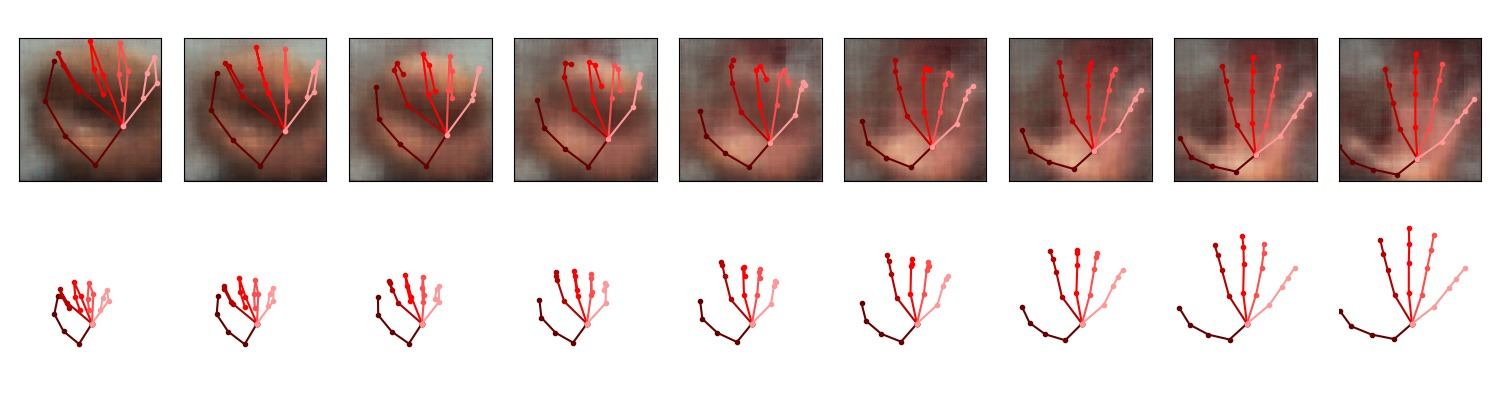}
\includegraphics[width=0.95\linewidth,clip,trim={0 13mm 0 58mm}]{./figures/RGB_3D_2D_interpolation/figure_img2598_to_img2870.png}
\includegraphics[width=0.95\linewidth,clip,trim={0 52mm 0 8mm}]{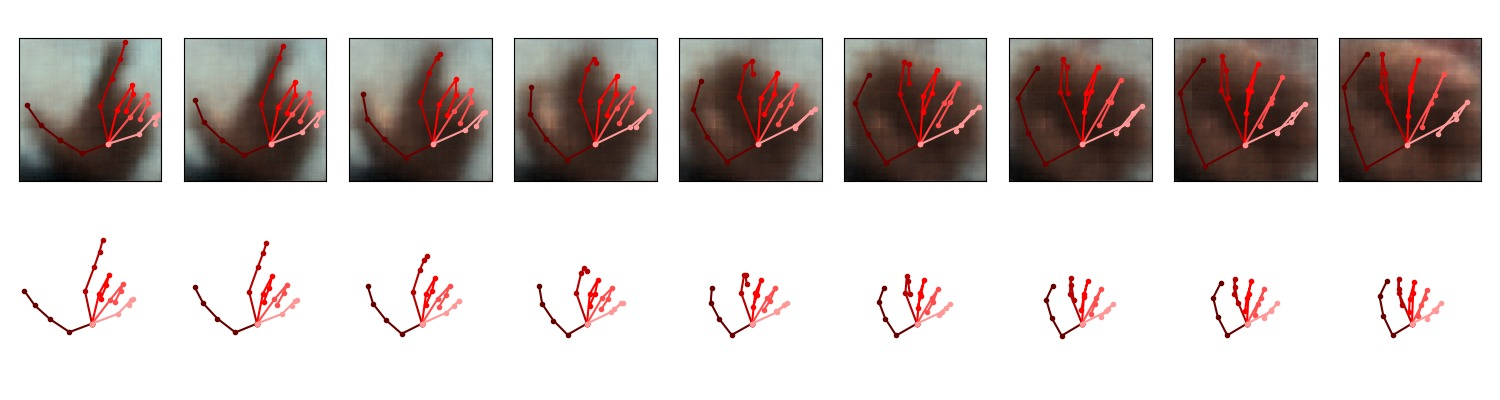}
\includegraphics[width=0.95\linewidth,clip,trim={0 13mm 0 58mm}]{./figures/RGB_3D_2D_interpolation/figure_img14366_to_img14656.png}
\includegraphics[width=0.95\linewidth,clip,trim={0 52mm 0 8mm}]{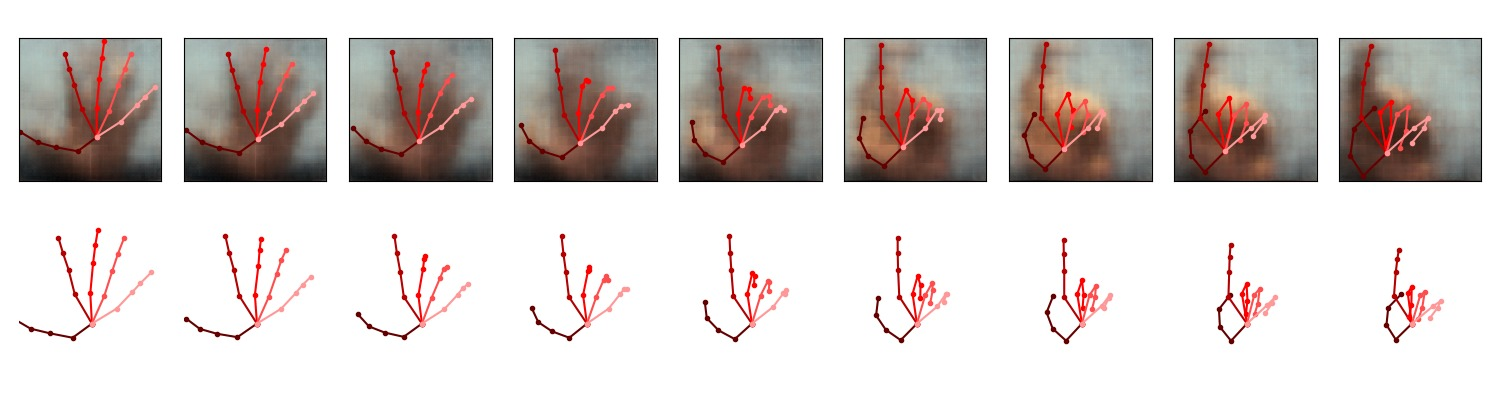}
\includegraphics[width=0.95\linewidth,clip,trim={0 13mm 0 58mm}]{./figures/RGB_3D_2D_interpolation/figure_img18180_to_img18482.png}
\end{center}
\end{subfigure}
\caption{\textbf{Latent space walk}. The left-most and right-most figures are reconstruction from latent space samples of two real RGB images. The figures in-between are multi-modal reconstruction from interpolated latent space samples, hence are completely synthetic. Shown are the reconstructed RGB images, with the reconstructed 2D keypoints (overlayed on the RGB image) and the corresponding reconstructed 3D joint skeleton. Each column-triplet is created from the same point in the latent space.}
\label{fig:latent_space_walk_2D3DRGB}
\end{figure*}

To further evaluate this property, in Fig.~\ref{fig:latent_space_walk_2D3DRGB} we show samples from the manifold, decoding them into different modalities. The latent samples are chosen such that the lie on an interpolated line between two embedded images. In other words, we took sample $x^1_{RGB}$ and $x^2_{RGB}$ and encoded them to obtain latent sample $z^1$ and $z^2$. We then interpolated linearly between these two latent samples, obtaining latent samples $z^j$ which were then decoded into the 2D, 3D and RGB modality, resulting in a triplet. Hence the left-most and right-most samples of the figure correspond to reconstruction of the RGB image and prediction of its 2D and 3D keypoints, whereas the middle figures are completely synthetic. It's important to note here that each decoded triplet originates from the same point in the latent space. This visualization shows that our learned manifold is indeed consistent amongst all three modalities. This result is in-line with the visualization of the joint embedding space visualized in \figref{fig:tsne_embedding}.

\textbf{Additional figures.}
Fig. \ref{fig:res_rgb_real} visualizes predictions on STB. The poses contained in the dataset are simpler, hence the predictions are very accurate. Sometimes the estimated hand poses even appear to be more correct than the ground truth (cf. right most column).
Fig. \ref{fig:res_rgb_synth} shows predictions on RHD. The poses are considerably harder than in the STB dataset and contain more self-occlusion. Nevertheless, our model is capable of predicting realistic poses, even for occluded joints. Fig. \ref{fig:res_depth} shows similar results for depth images.

\begin{figure*}[h!]
\begin{center}
\includegraphics[width=0.95\linewidth,clip,trim={0 25mm 0 25mm}]{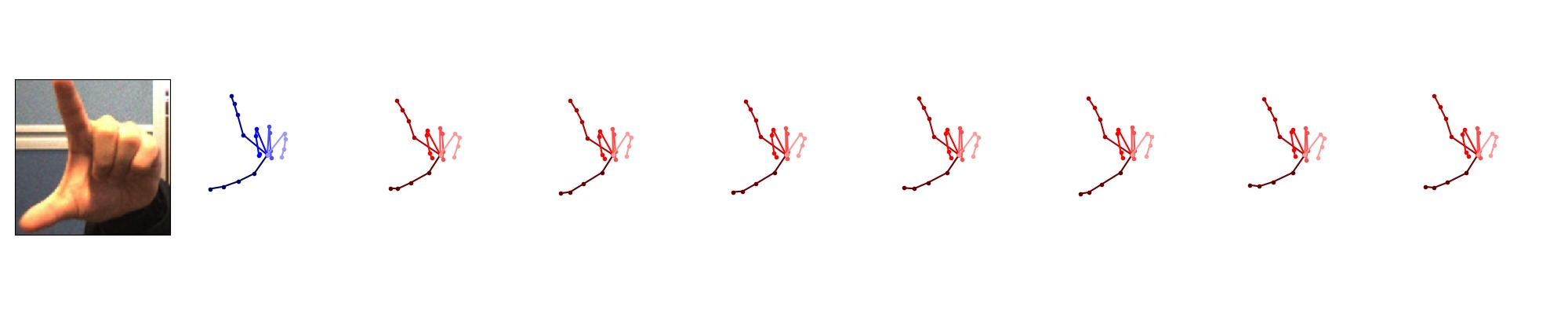}
\includegraphics[width=0.95\linewidth,clip,trim={0 25mm 0 25mm}]{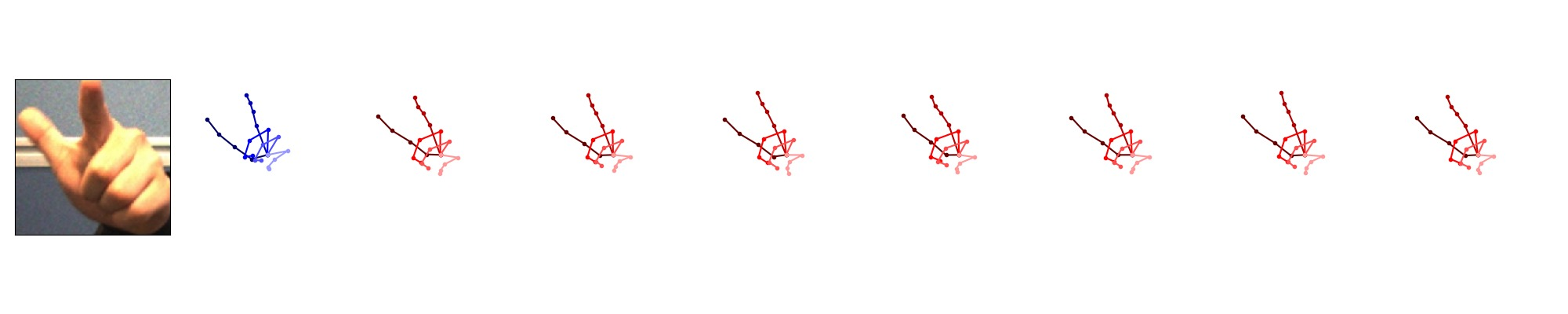}
\includegraphics[width=0.95\linewidth,clip,trim={0 25mm 0 25mm}]{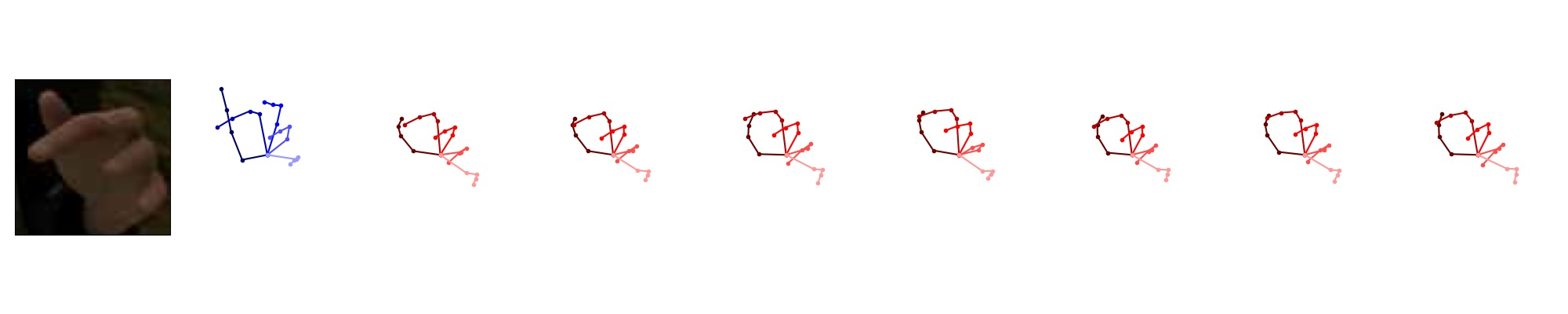}
\includegraphics[width=0.95\linewidth,clip,trim={0 25mm 0 25mm}]{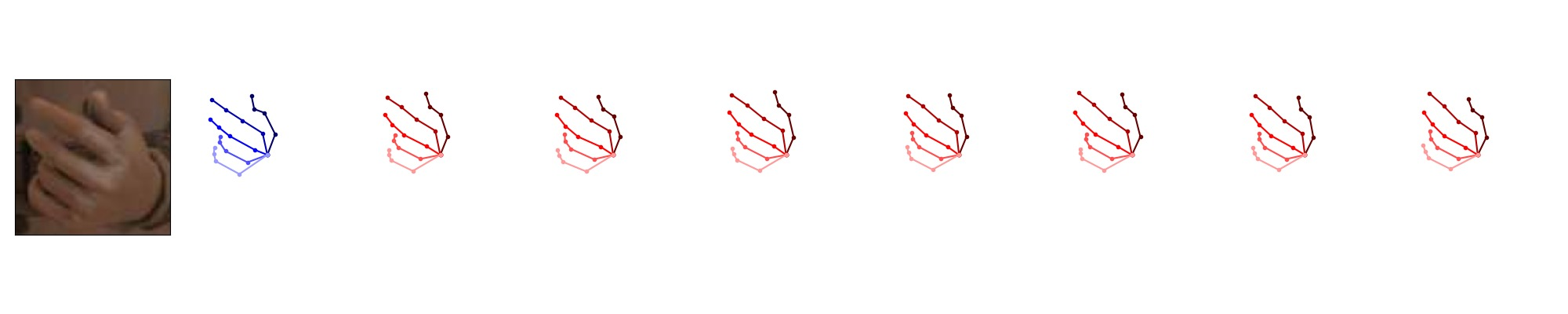}
\end{center}
\vspace*{-4mm}
\caption{\textbf{Sampling from prediction}. This figure shows the resulting reconstruction from samples $z\sim\mathcal{N}(\mu,\sigma^2)$ (red), where $\mu, \sigma^2$ are the predicted mean and variance output by the RGB encoder. Ground-truth is provided in blue for comparison.}
\label{fig:prediction_sampling}
\end{figure*}

Fig. \ref{fig:prediction_sampling} displays the input image, its ground truth joint skeleton and predictions of our model. These were constructed by sampling repeatedly from the latent space from the predicted mean and variance which are produced by the RGB encoder. Generally, there are only minor variations in the pose, showing the high confidence of predictions of our model.


\subsection{Influence of model capacity}
All of our models predicting 3D joint skeleton from RGB images have strictly less parameters than \cite{zimmermann2017}. Our smallest model consists of $12'398'387$ parameters, and the biggest ranges up to $14'347'346$. In comparison, \cite{zimmermann2017} uses $21'394'529$ parameters. Yet, we still outperform them on RHD and reach parity on the saturated STB dataset. This provides further evidence of the proposed approach to learn a manifold of physically plausible hand configurations and to leverage this for the prediction of joint positions directly from an RGB image.\newline
\cite{oberweger2017} employ a ResNet-50 architecture to predict the 3D joint coordinates directly from depth. In the experiment reported in the main paper, our architecture produced a slightly higher mean EPE (8.5) in comparison to DeepPrior++ (8.1). We believe this can be mostly attributed to differences in model capacity. To show this, we re-ran our experiment on depth images, using the ResNet-50 architecture as encoder and achieved a mean EPE of 8.0.

\begin{figure*}[th!]
\newcommand{\stbvis}[3]{
  \includegraphics[trim={12mm 32mm 252mm 28mm},clip,width=0.10\linewidth]{#1}
  \includegraphics[trim={{\dimexpr 102mm-#2mm\relax} {\dimexpr 32mm-#3mm\relax} {\dimexpr 162mm+#2mm\relax} {\dimexpr 28mm+#3mm\relax}},clip,width=0.10\linewidth]{#1}
  \includegraphics[trim={{\dimexpr 224mm-#2mm\relax} {\dimexpr 32mm-#3mm\relax} {\dimexpr 40mm+#2mm\relax} {\dimexpr 28mm+#3mm\relax}},clip,width=0.10\linewidth]{#1}
}
\newcommand{\rhdvis}[3]{
  \includegraphics[trim={12mm 32mm 252mm 28mm},clip,width=0.10\linewidth]{#1}
  \includegraphics[trim={{\dimexpr 102mm-#2mm\relax} {\dimexpr 32mm-#3mm\relax} {\dimexpr 162mm+#2mm\relax} {\dimexpr 28mm+#3mm\relax}},clip,width=0.10\linewidth]{#1}
  \includegraphics[trim={{\dimexpr 224mm-#2mm\relax} {\dimexpr 32mm-#3mm\relax} {\dimexpr 40mm+#2mm\relax} {\dimexpr 28mm+#3mm\relax}},clip,width=0.10\linewidth]{#1}
}
\begin{subfigure}[t]{\linewidth}
\begin{center}
	\stbvis{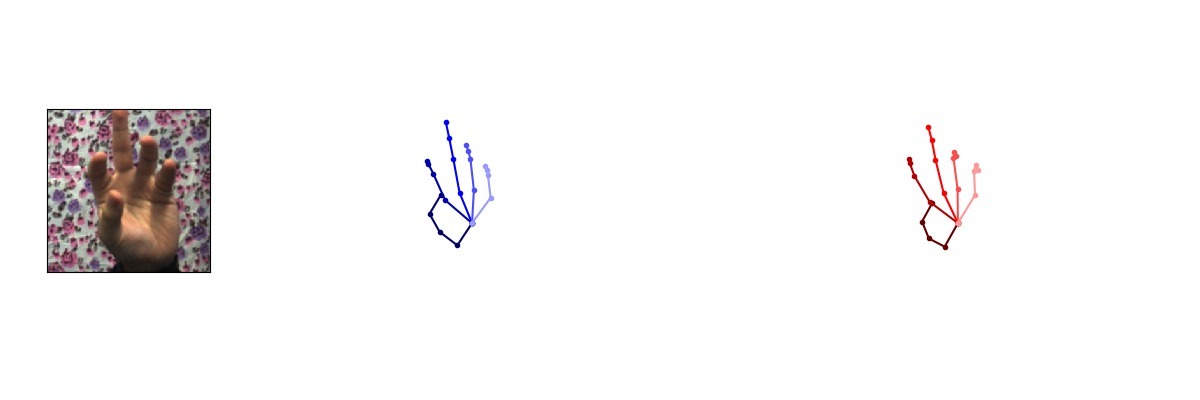}{6}{0}
	\stbvis{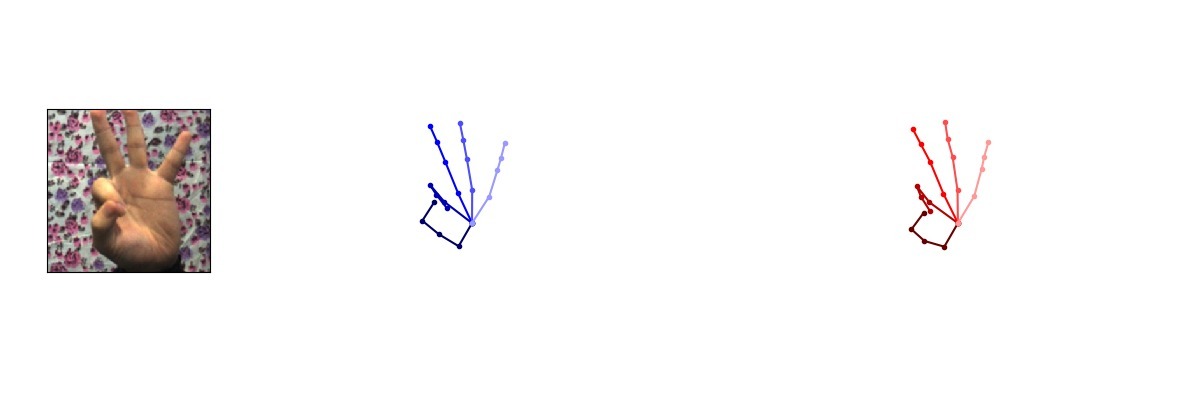}{8}{2}
	\stbvis{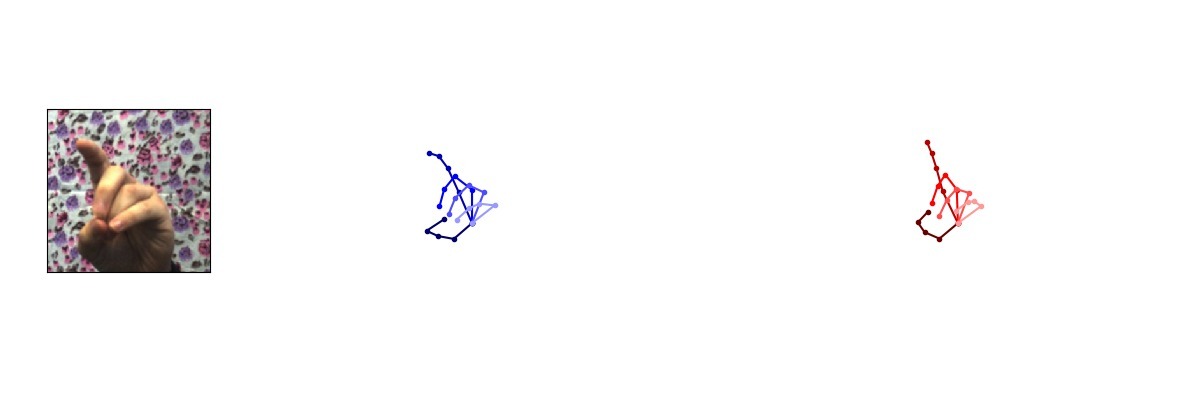}{6}{2} 
	\stbvis{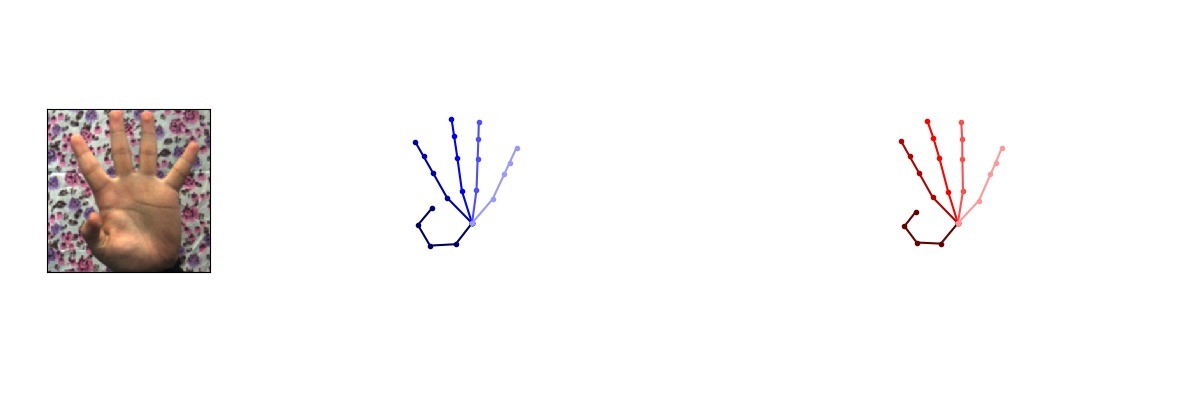}{6}{0}
	\stbvis{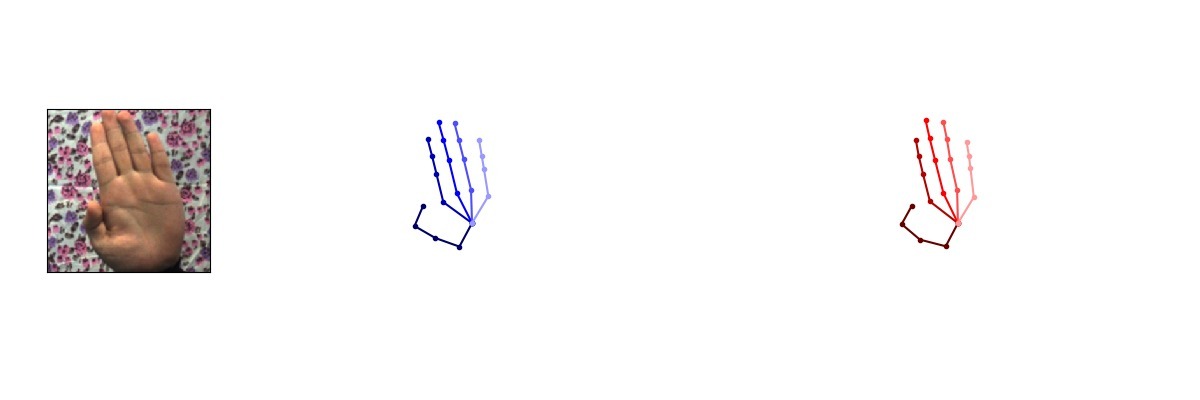}{6}{0} 
	\stbvis{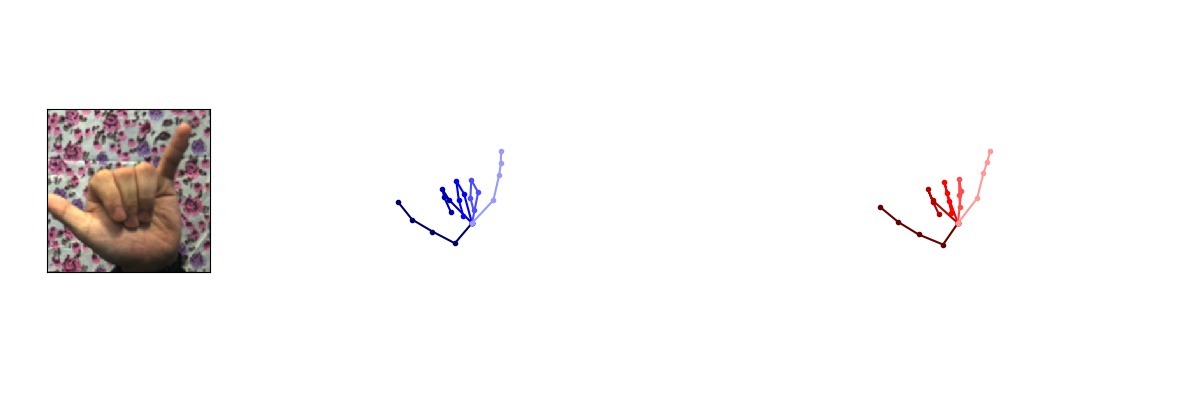}{3}{4}
	\stbvis{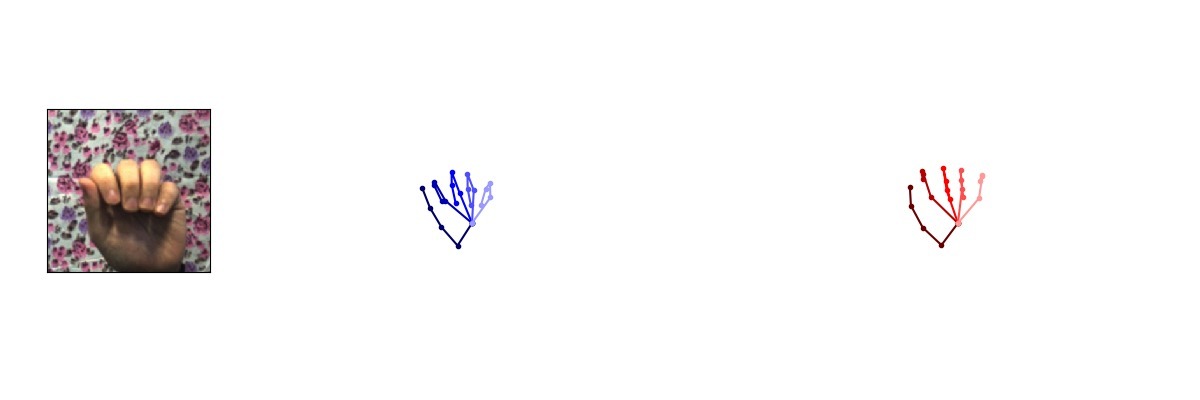}{6}{0}
	\stbvis{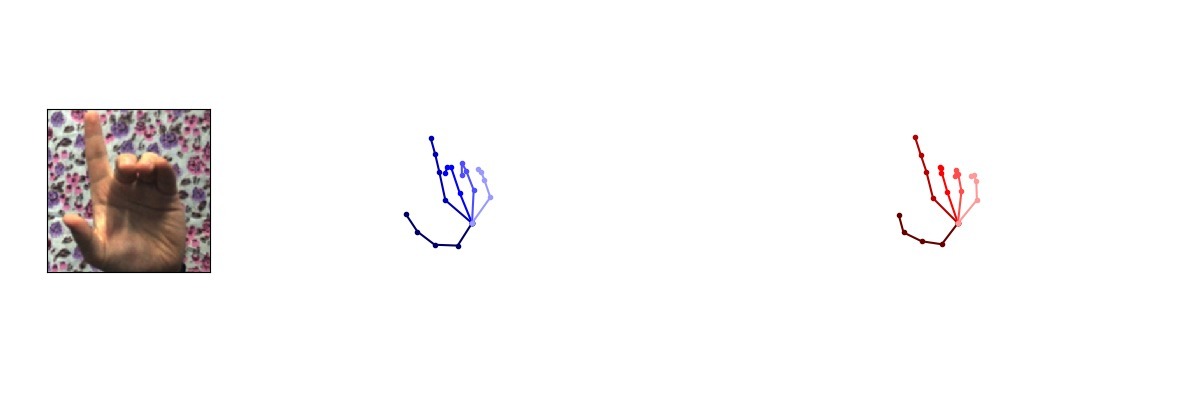}{6}{4}
	\stbvis{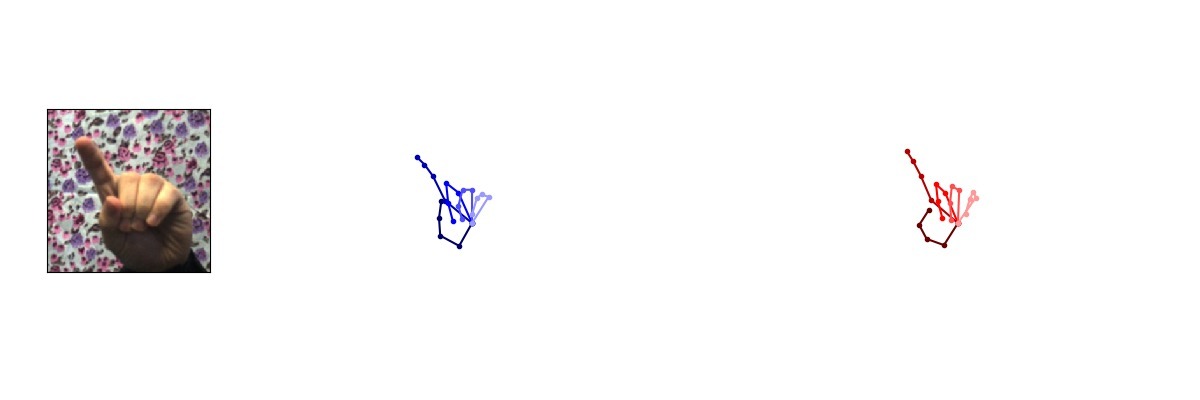}{3}{4}
	\stbvis{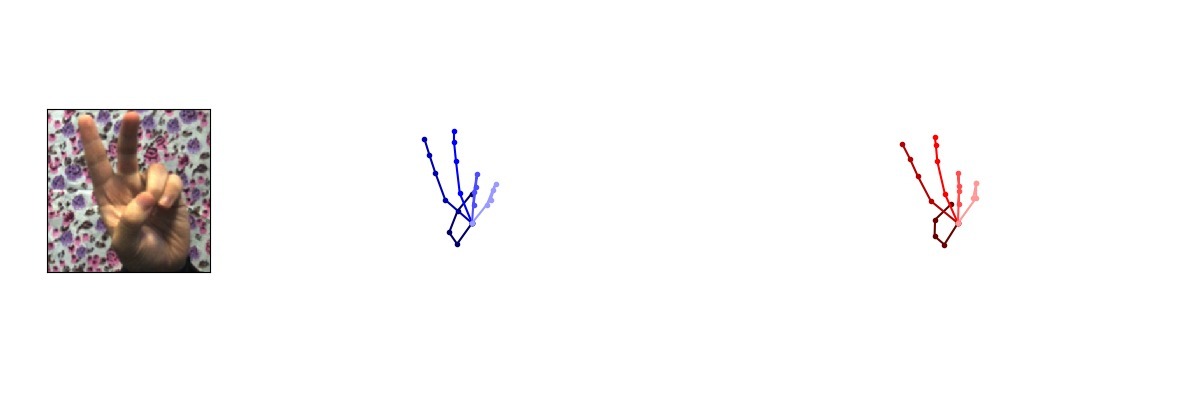}{6}{0}
	\stbvis{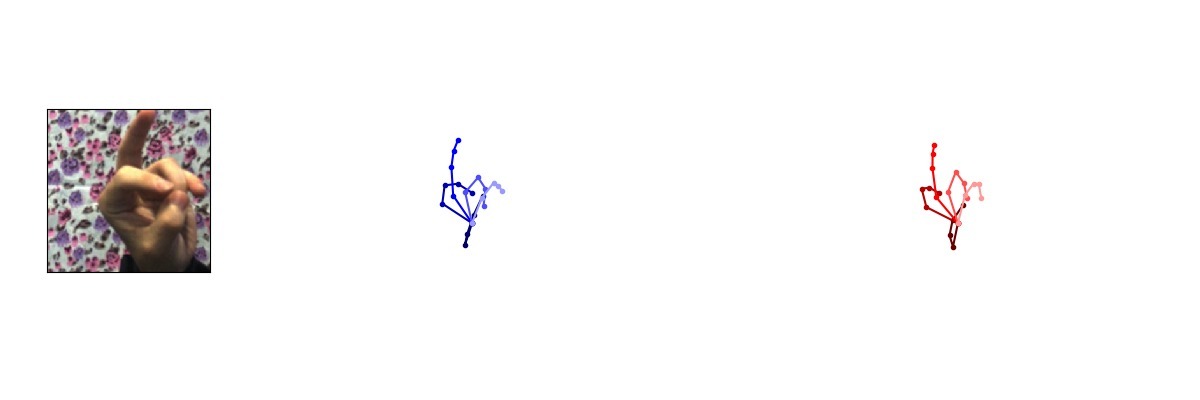}{6}{4}
	\stbvis{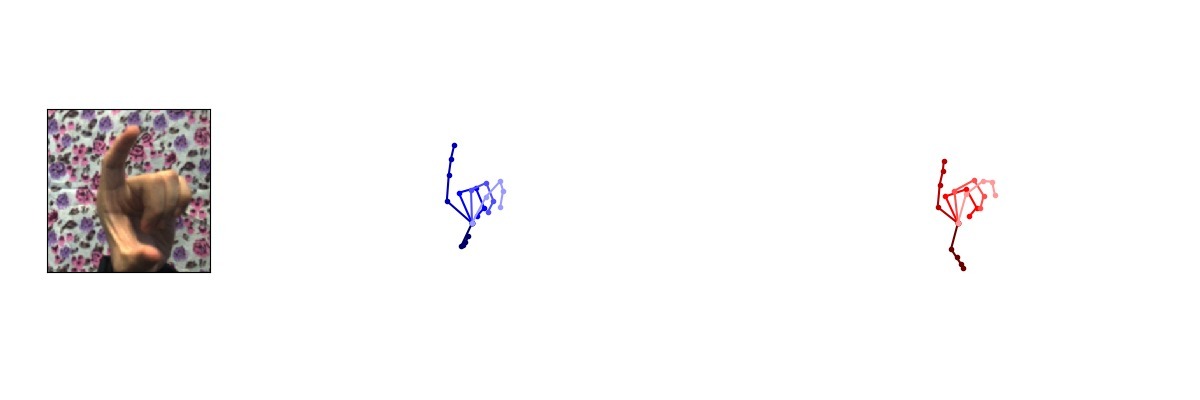}{3}{4}
	\stbvis{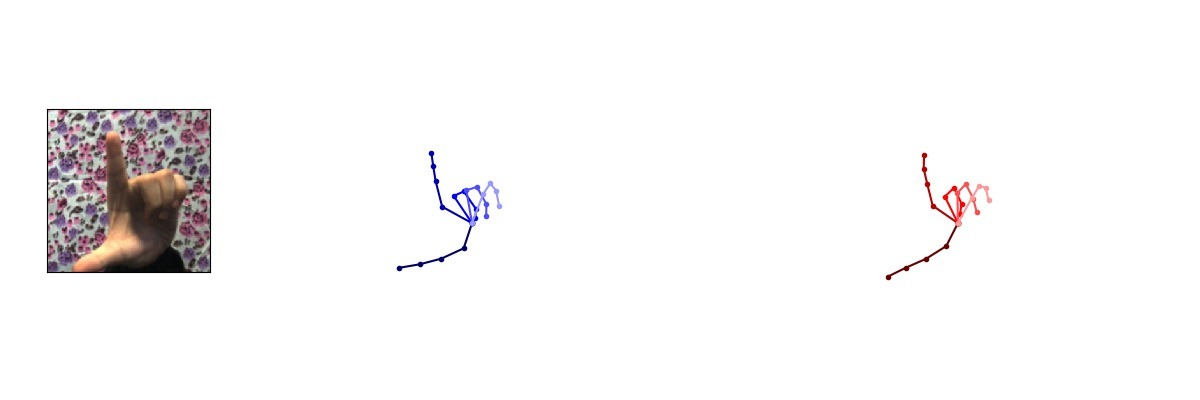}{6}{4}
	\stbvis{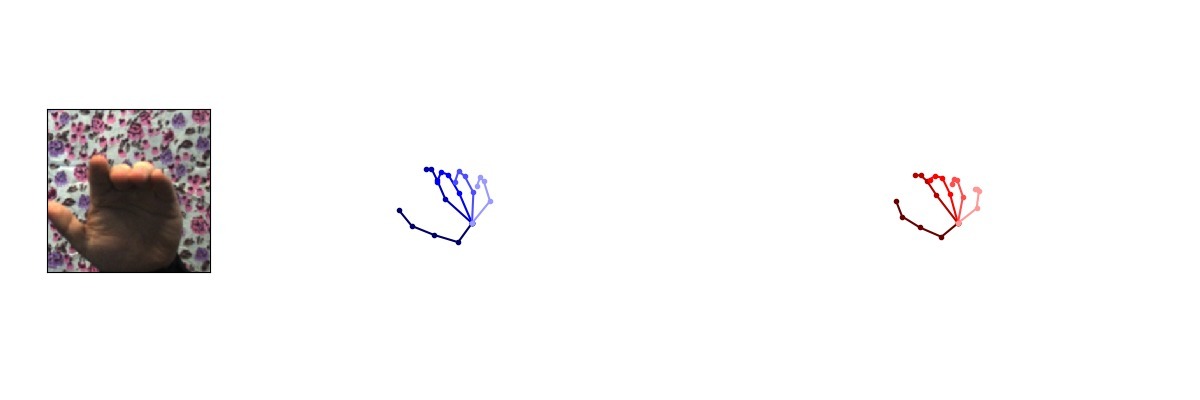}{6}{8}
	\stbvis{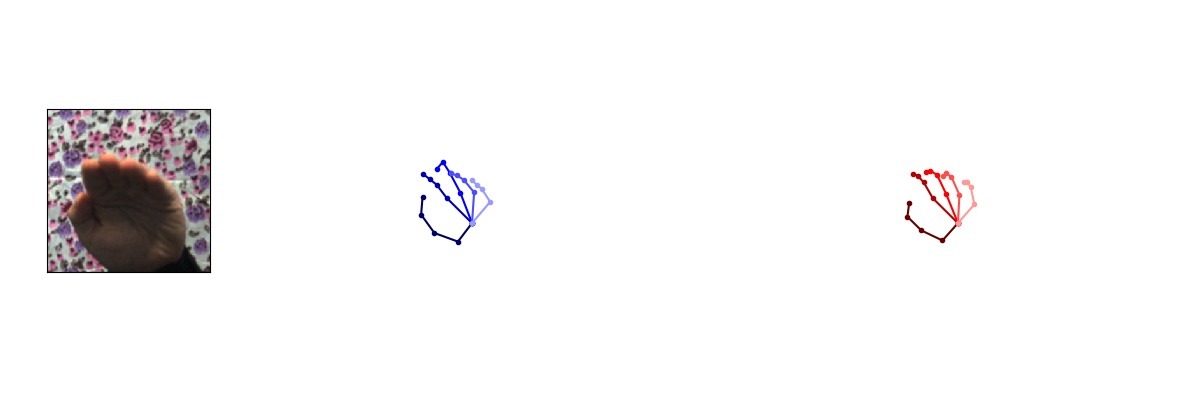}{3}{4}
\end{center}
\caption{\textbf{STB} (from RGB)}
\label{fig:res_rgb_real}
\end{subfigure}

\begin{subfigure}[t]{\linewidth}
\begin{center}
	\rhdvis{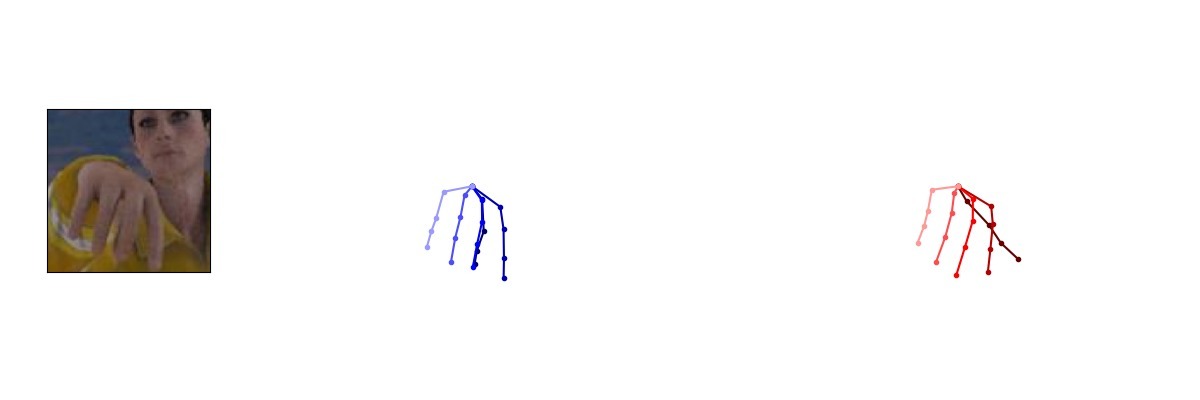}{-4}{2}
	\rhdvis{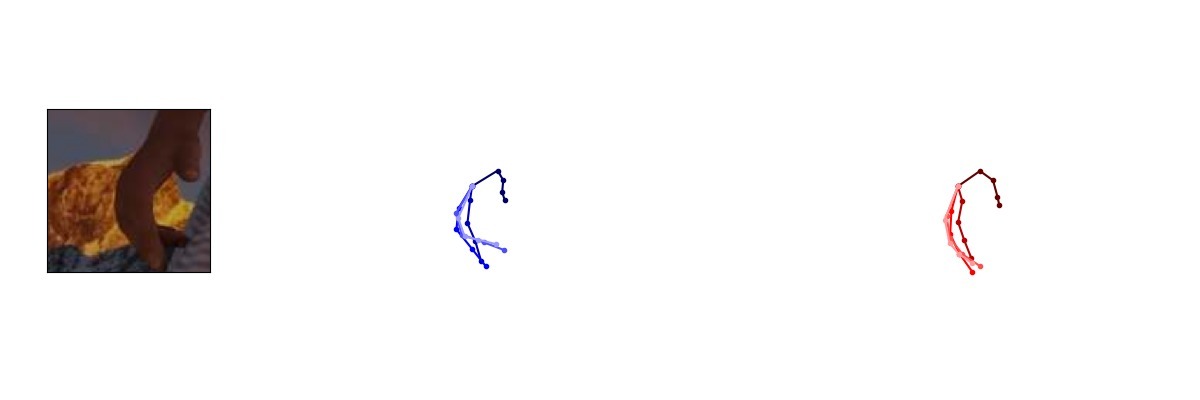}{4}{8}
	\rhdvis{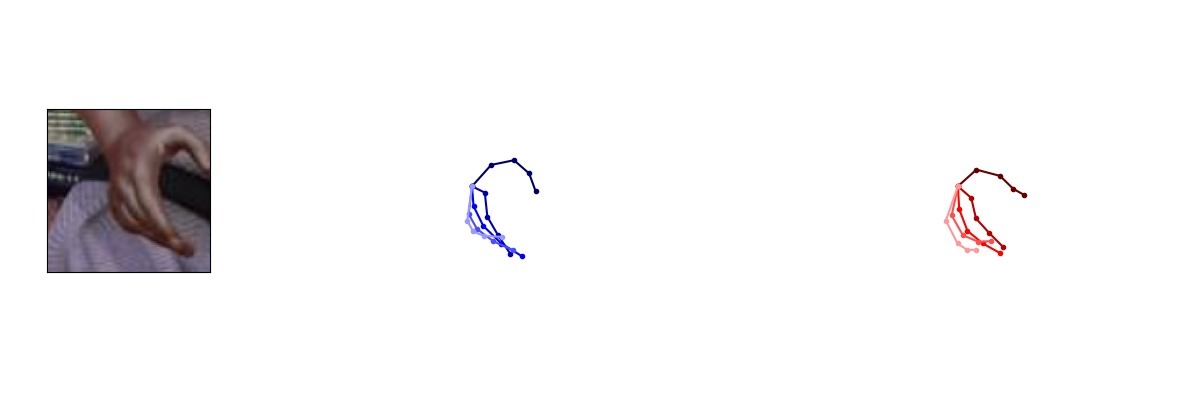}{-6}{2}
	\rhdvis{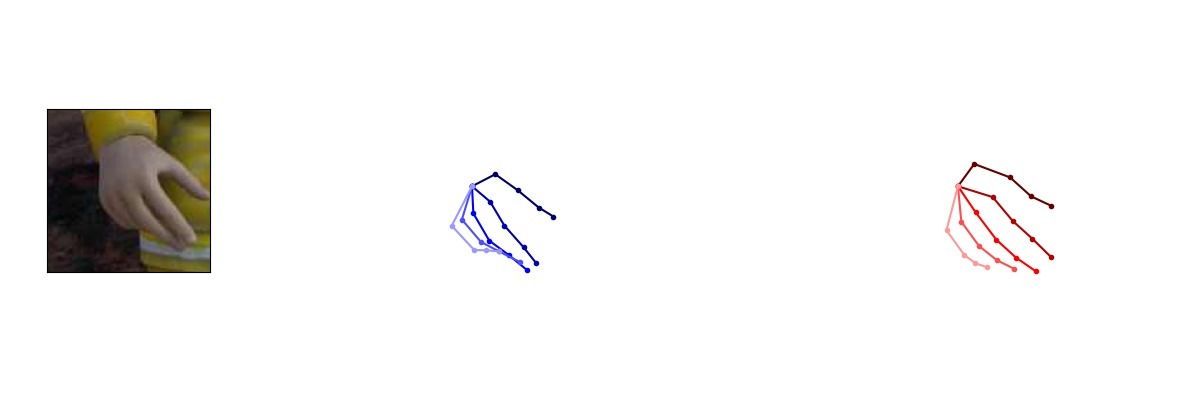}{-8}{8}
	\rhdvis{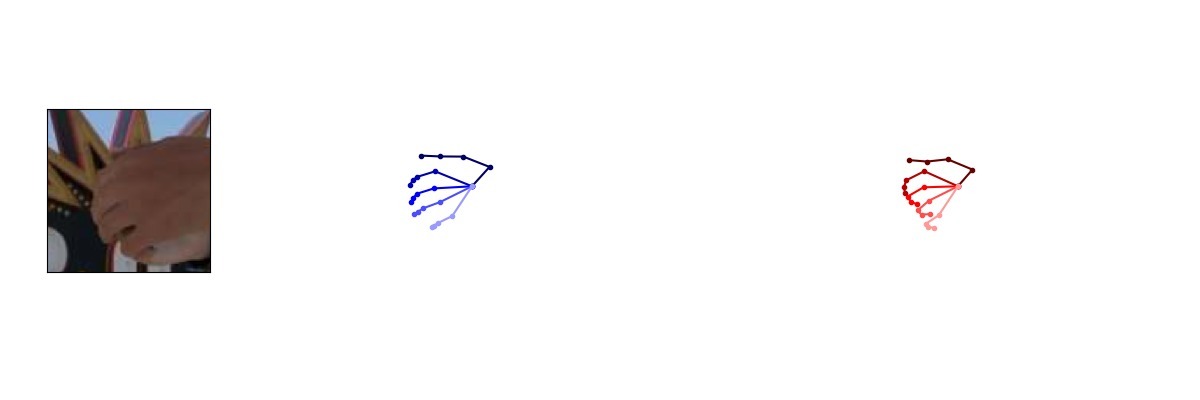}{8}{-4}
	\rhdvis{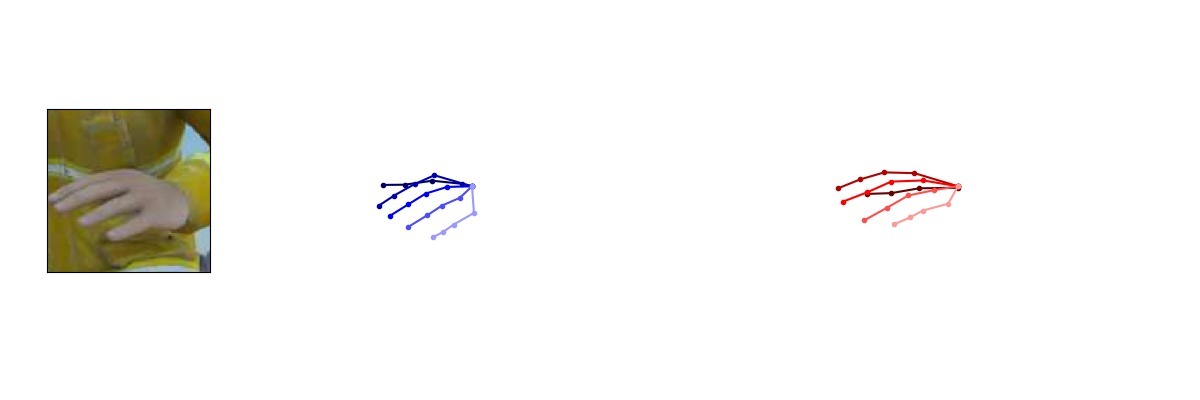}{12}{2}
	\rhdvis{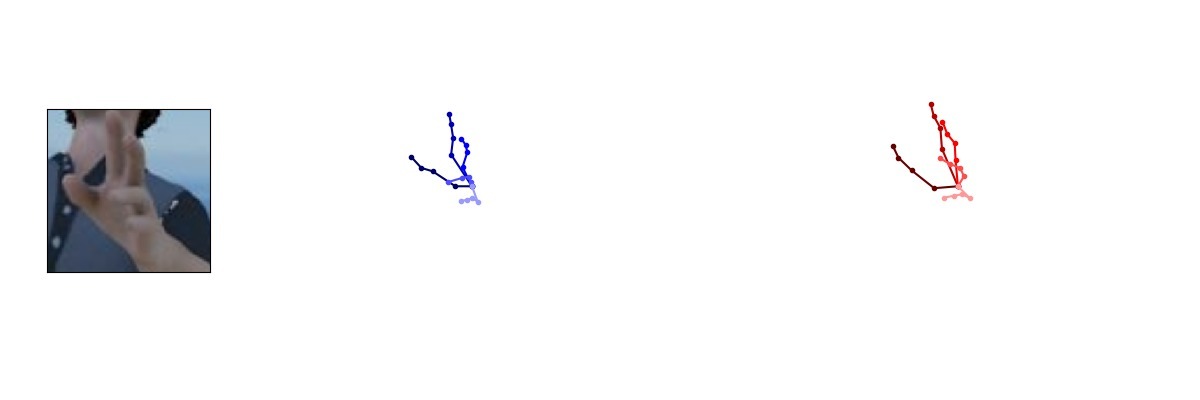}{5}{-5}
	\rhdvis{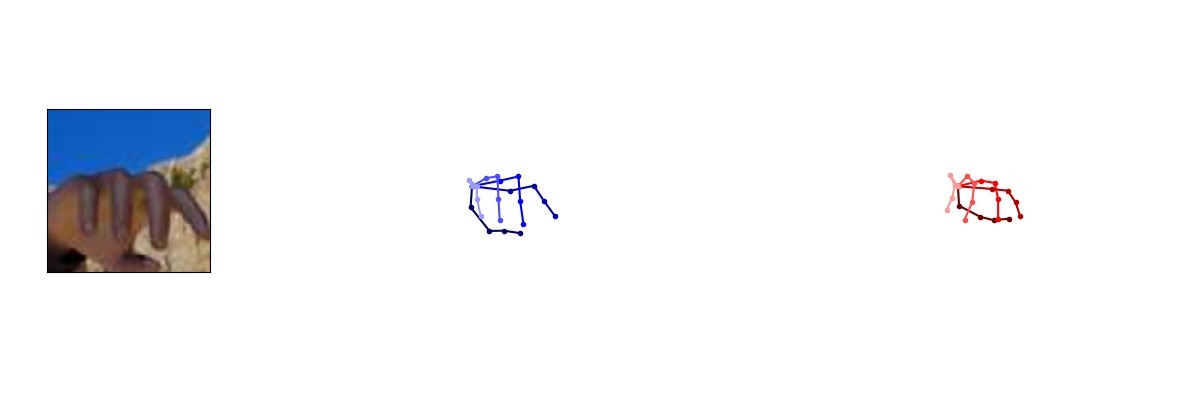}{0}{4}
	\rhdvis{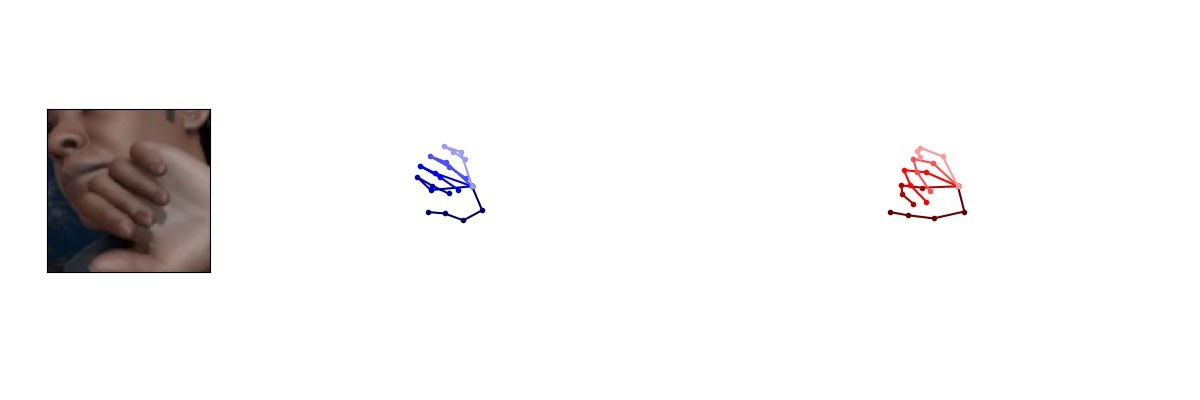}{8}{-2}
	\rhdvis{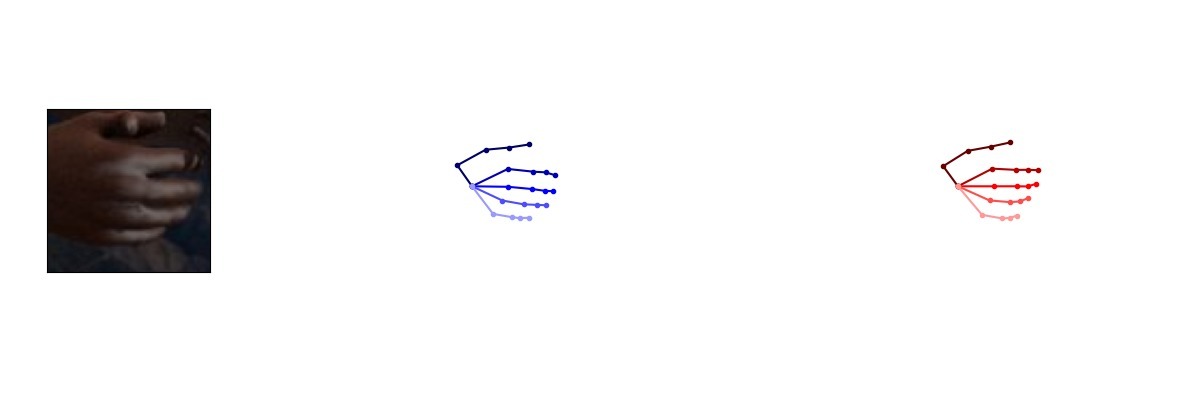}{-4}{0}
	\rhdvis{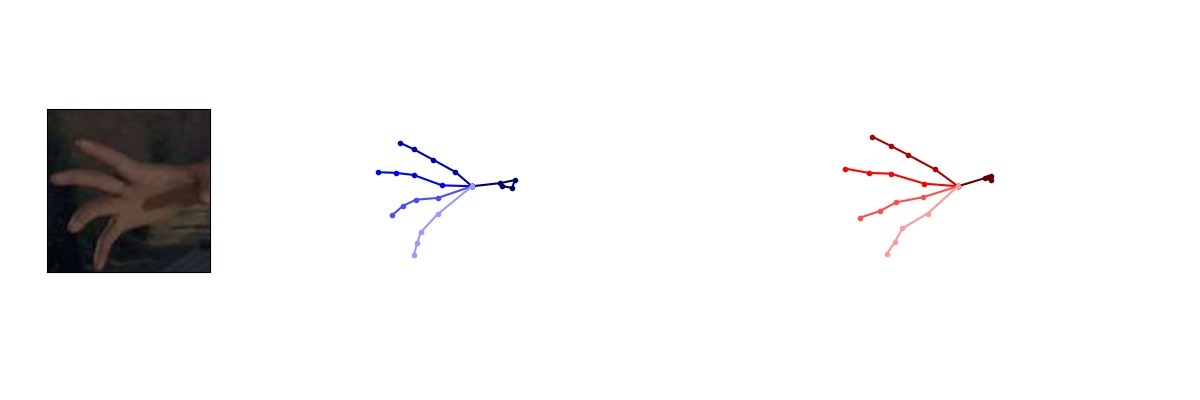}{8}{0}
	\rhdvis{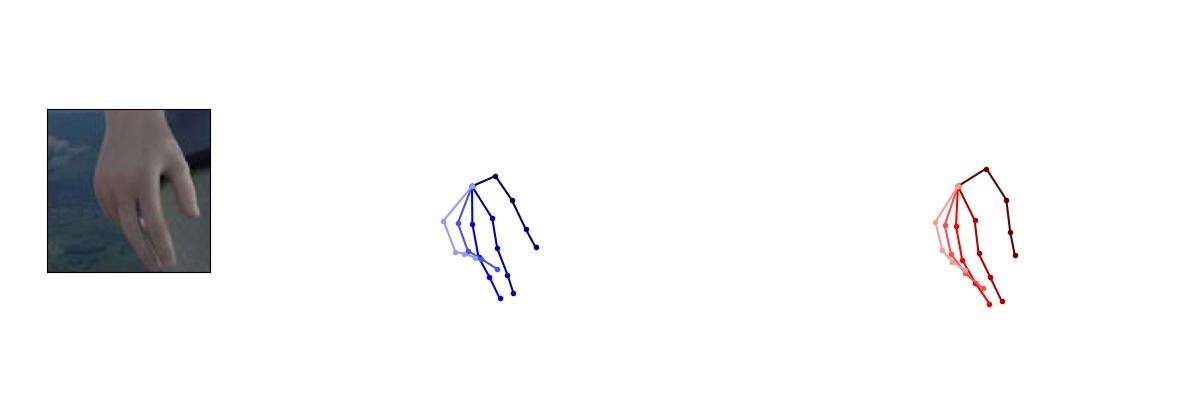}{4}{12}
	\rhdvis{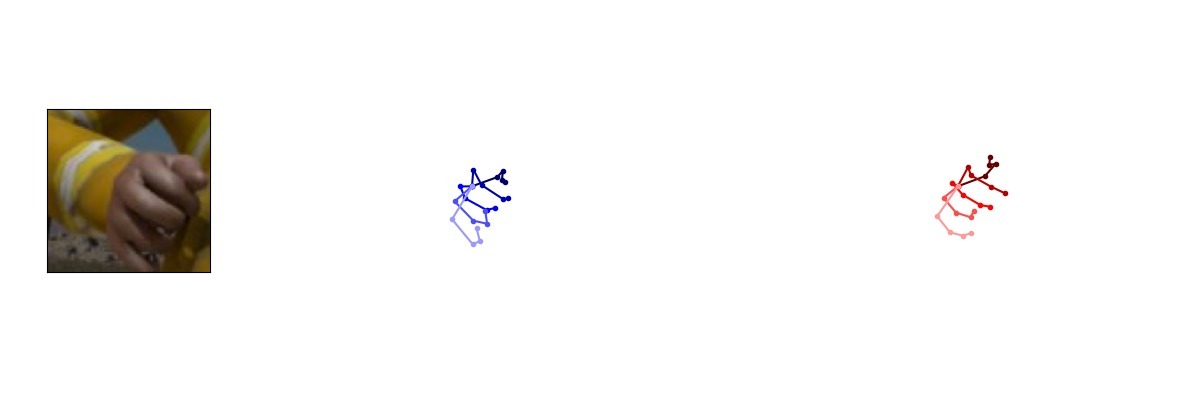}{0}{0}
	\rhdvis{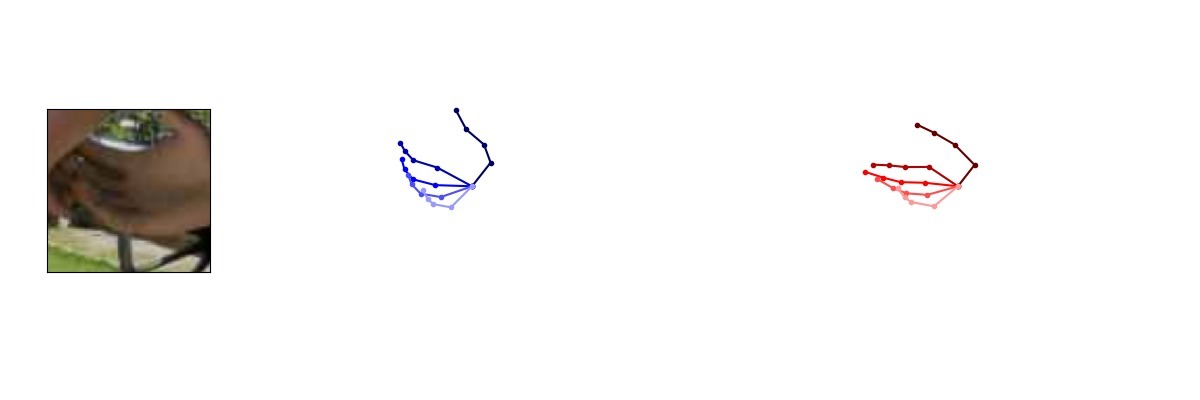}{8}{-5}
	\rhdvis{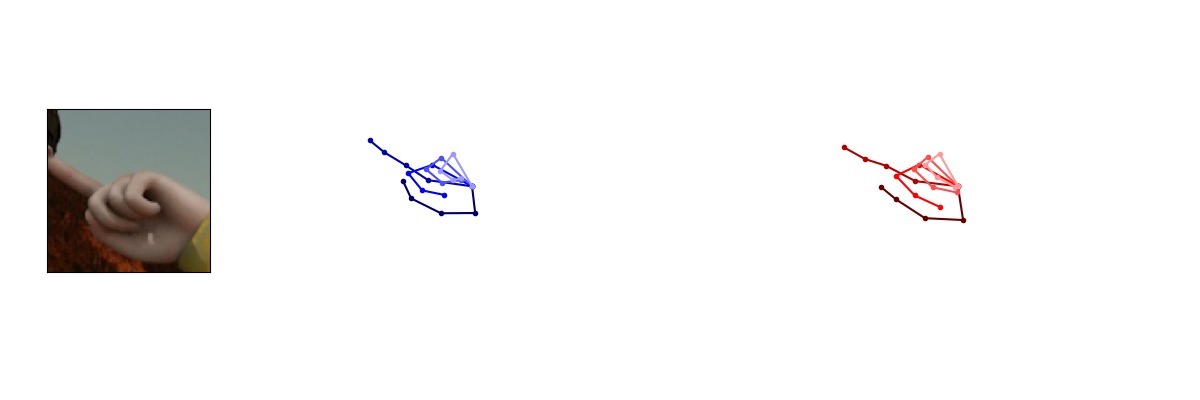}{12}{-4}
\end{center}
\caption{\textbf{RHD} (from RGB)}
\label{fig:res_rgb_synth}
\end{subfigure}
\caption{\textbf{RGB to 3D joint prediction}. Blue is ground truth and red is the prediction of our model.}
\end{figure*}

\begin{figure*}
\centering
\includegraphics[width=0.32\linewidth,clip,trim={0 18mm 0 8mm}]{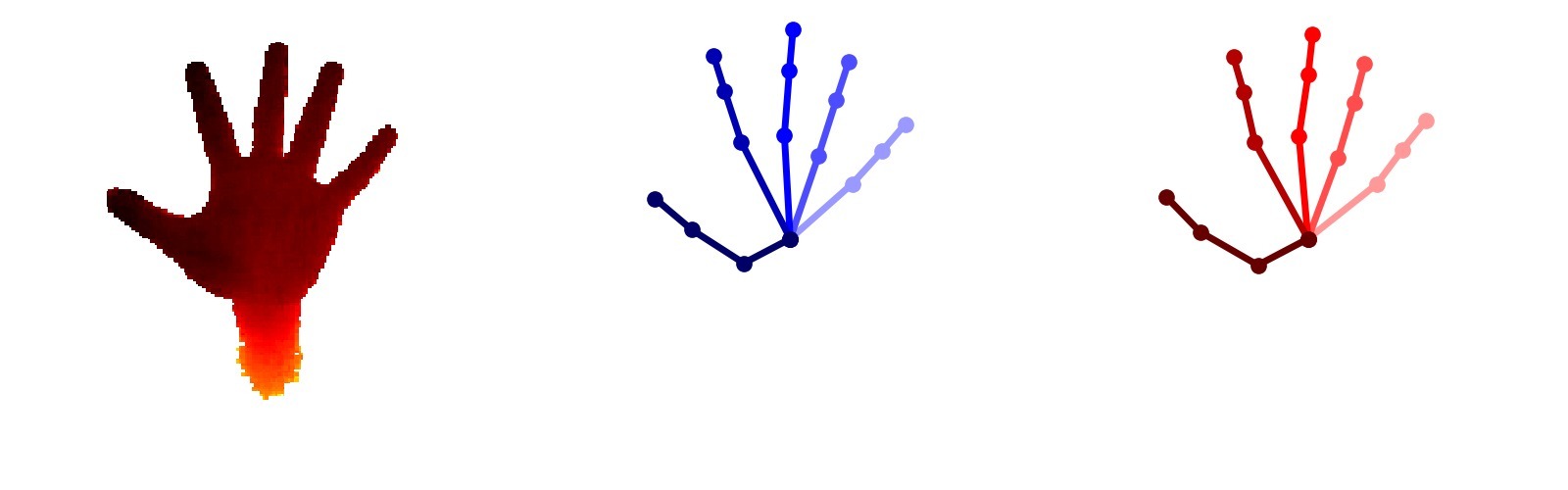}
\includegraphics[width=0.32\linewidth,clip,trim={0 18mm 0 8mm}]{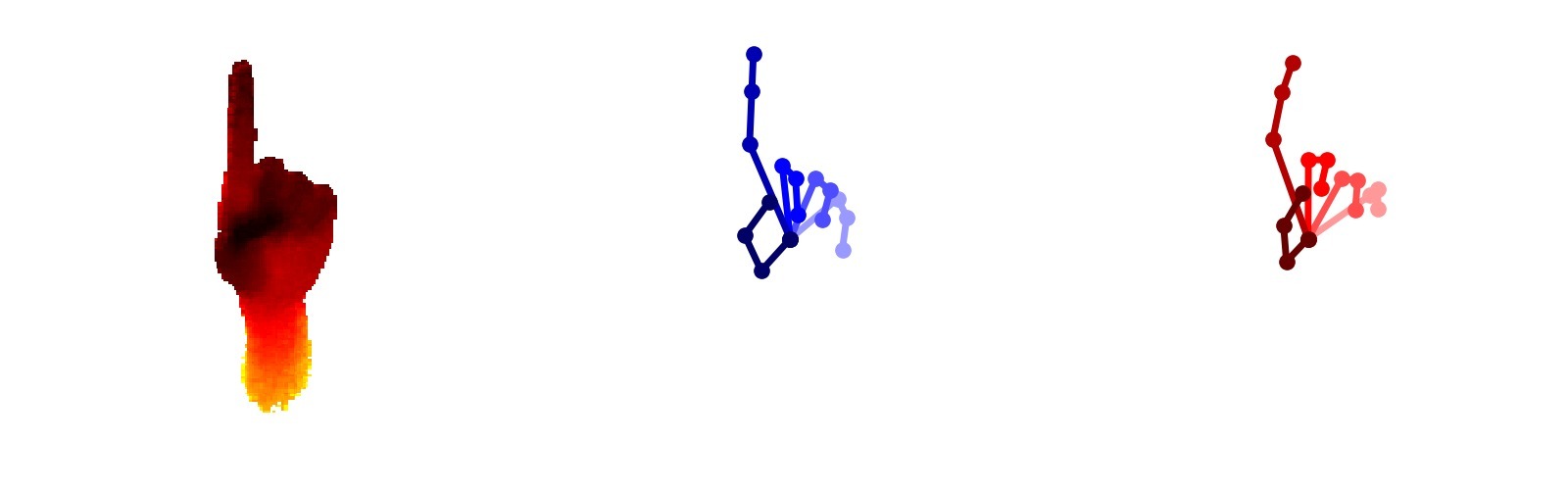}
\includegraphics[width=0.32\linewidth,clip,trim={0 18mm 0 8mm}]{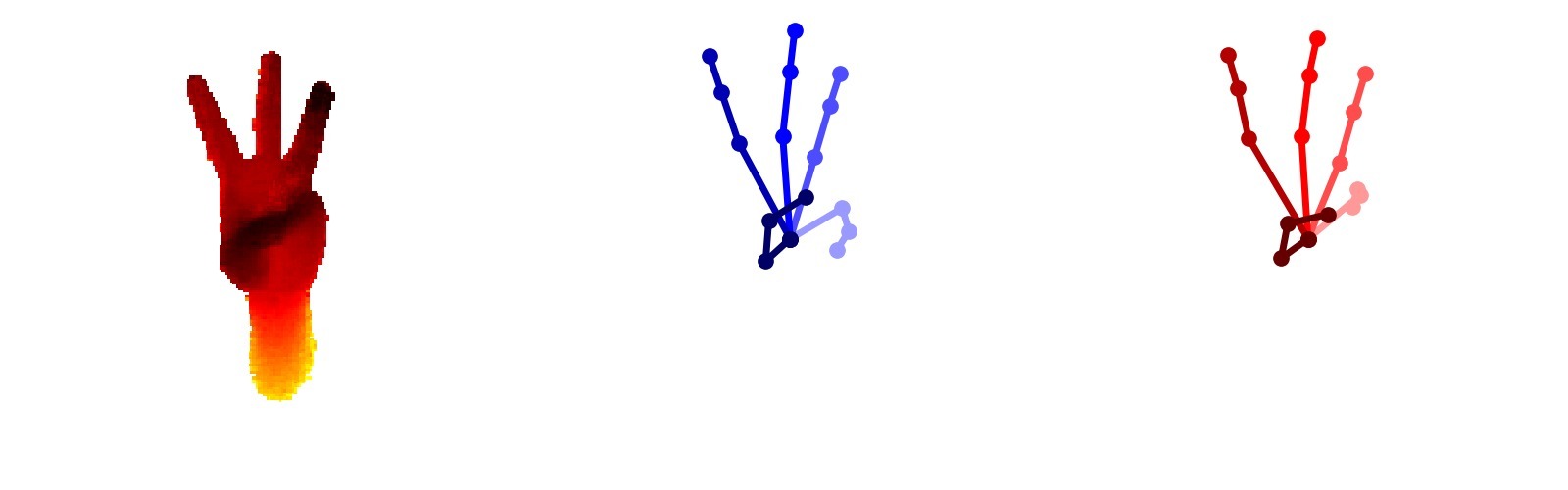}
\includegraphics[width=0.32\linewidth,clip,trim={0 18mm 0 8mm}]{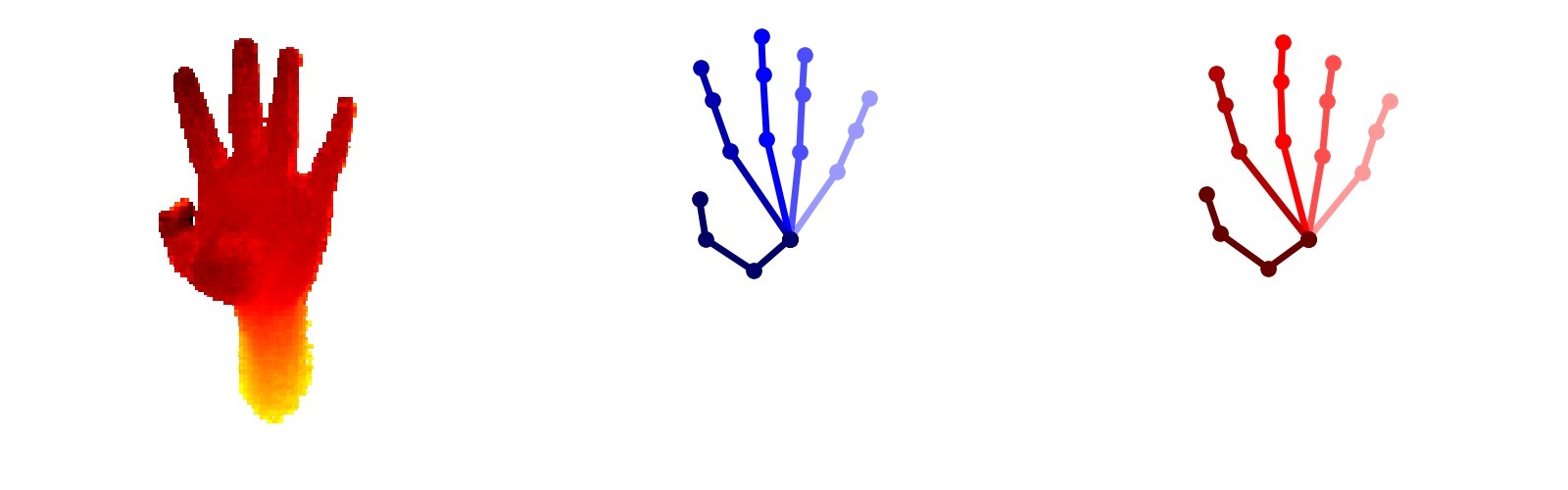}
\includegraphics[width=0.32\linewidth,clip,trim={0 18mm 0 8mm}]{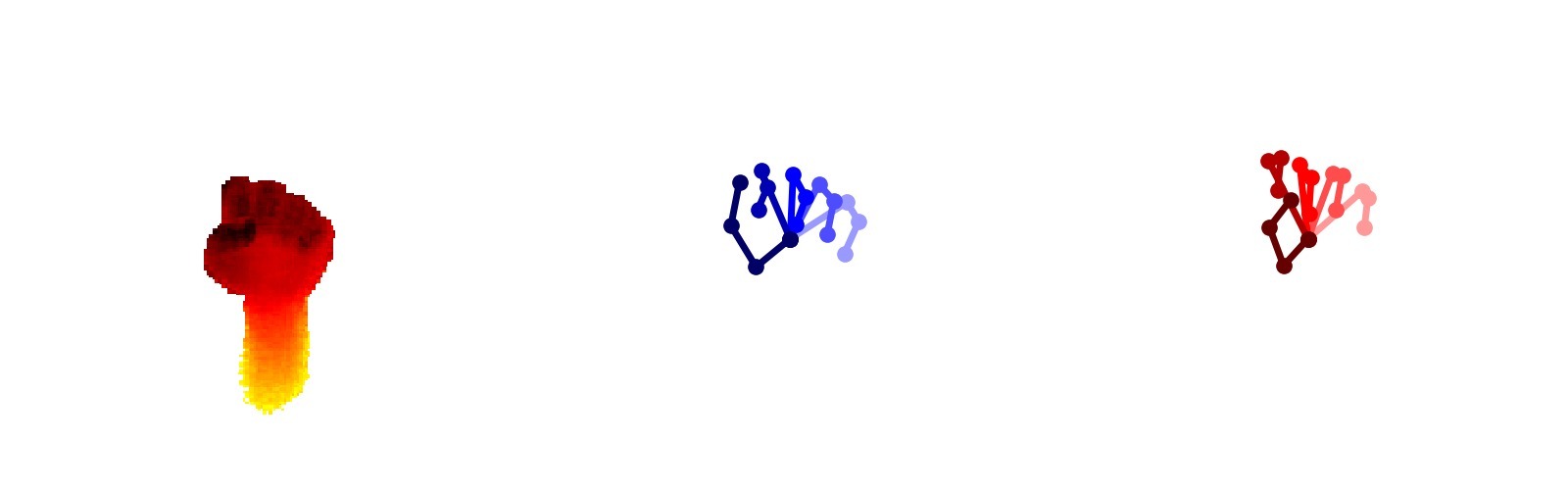}
\includegraphics[width=0.32\linewidth,clip,trim={0 18mm 0 8mm}]{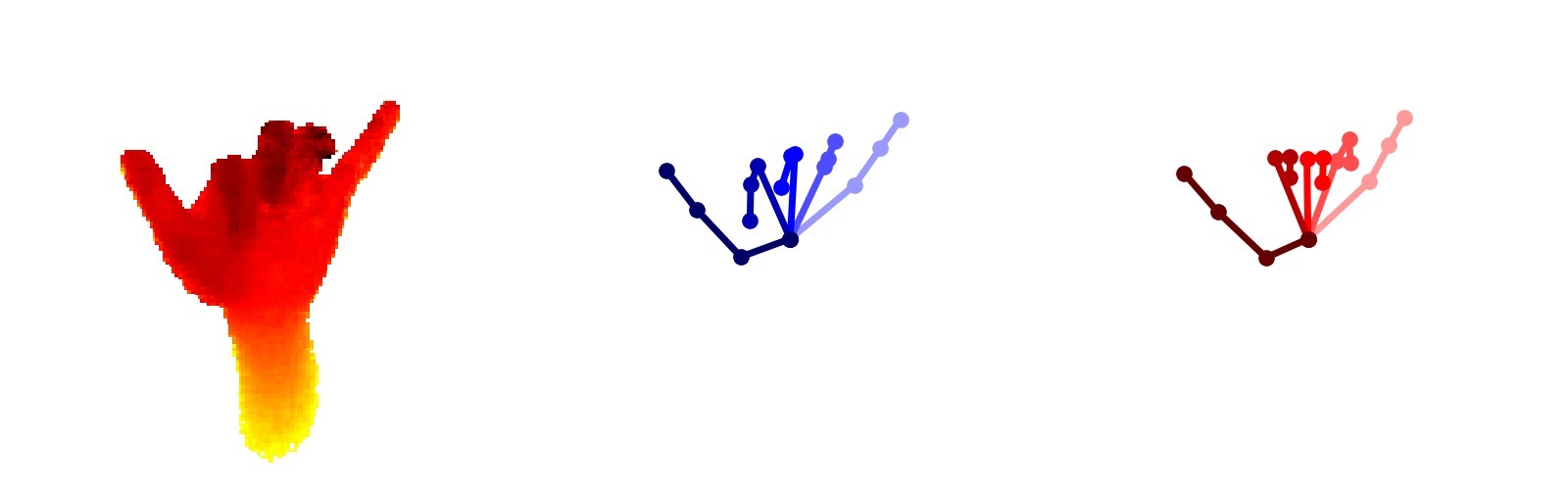}
\includegraphics[width=0.32\linewidth,clip,trim={0 18mm 0 8mm}]{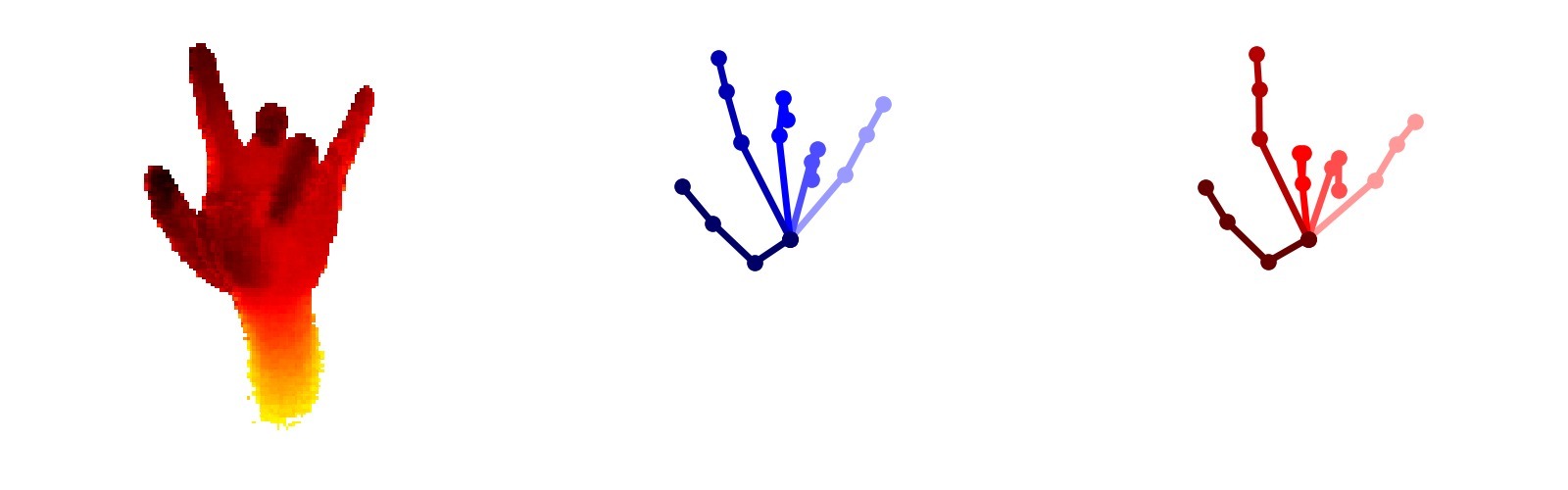}
\includegraphics[width=0.32\linewidth,clip,trim={0 18mm 0 8mm}]{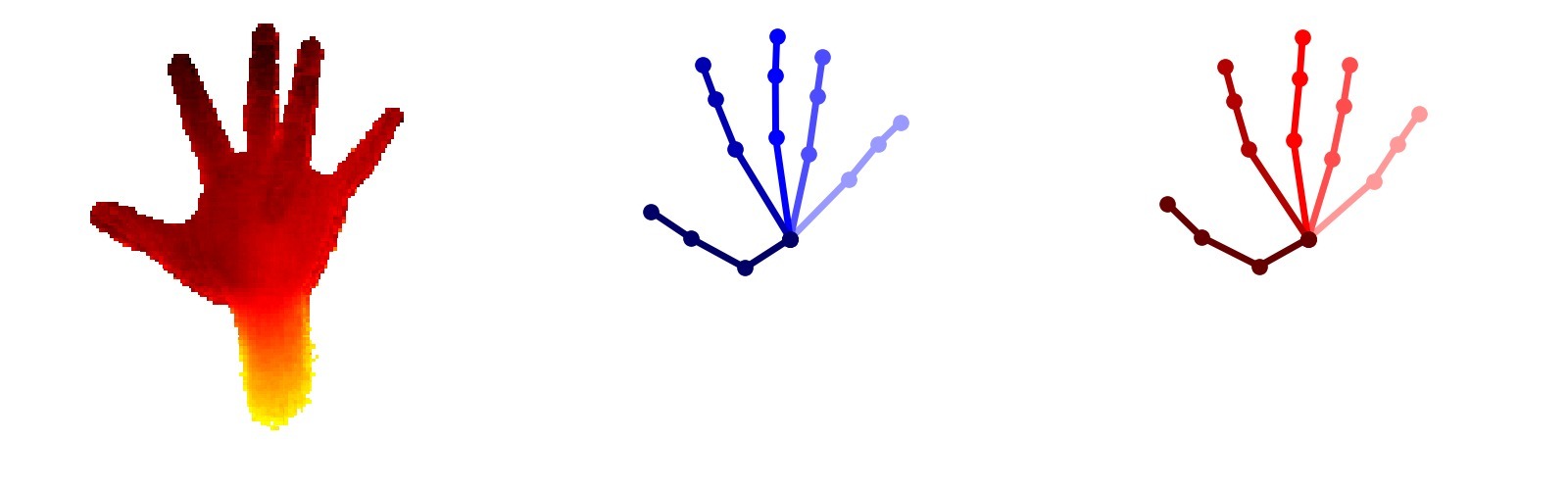}
\includegraphics[width=0.32\linewidth,clip,trim={0 18mm 0 8mm}]{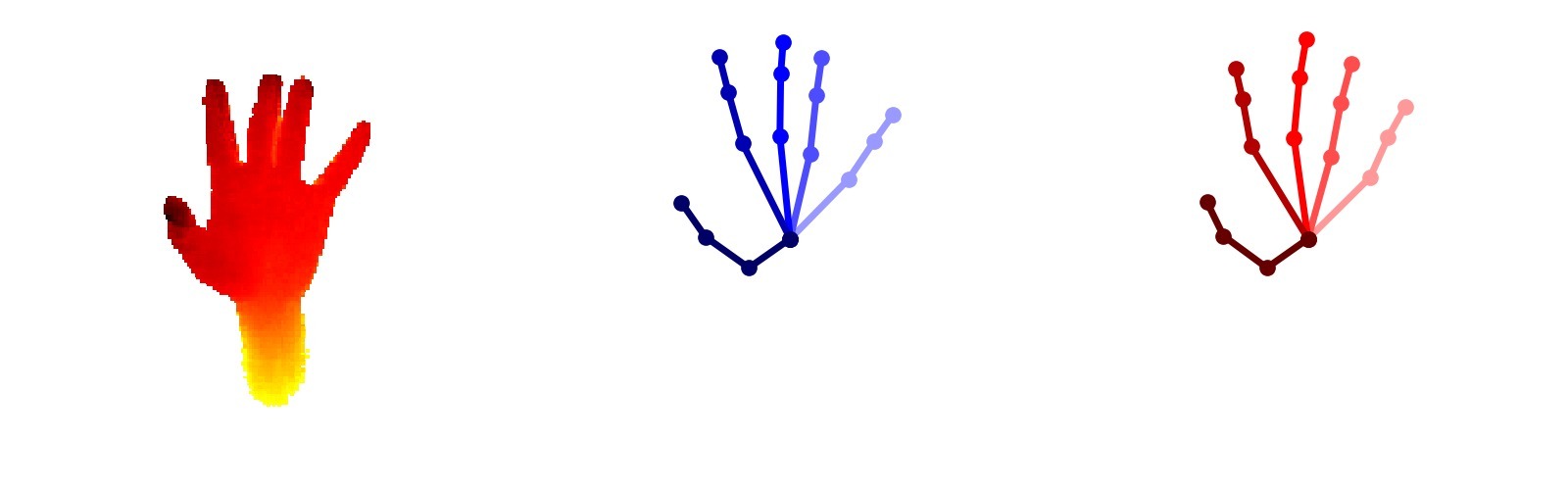}
\includegraphics[width=0.32\linewidth,clip,trim={0 18mm 0 8mm}]{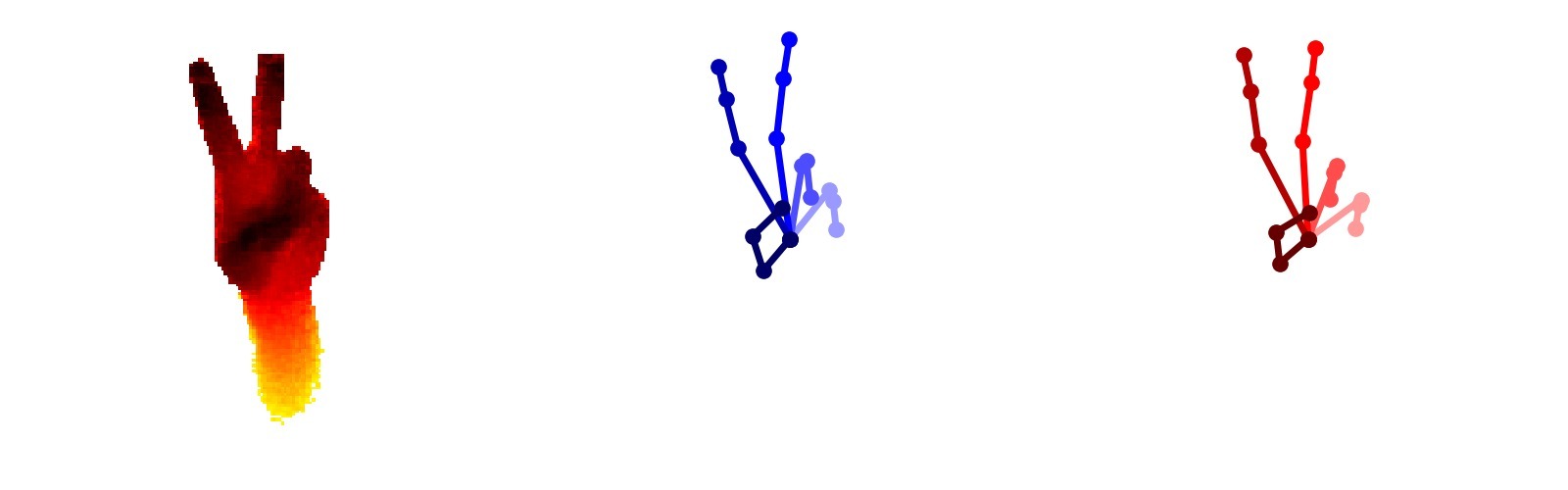}
\includegraphics[width=0.32\linewidth,clip,trim={0 18mm 0 8mm}]{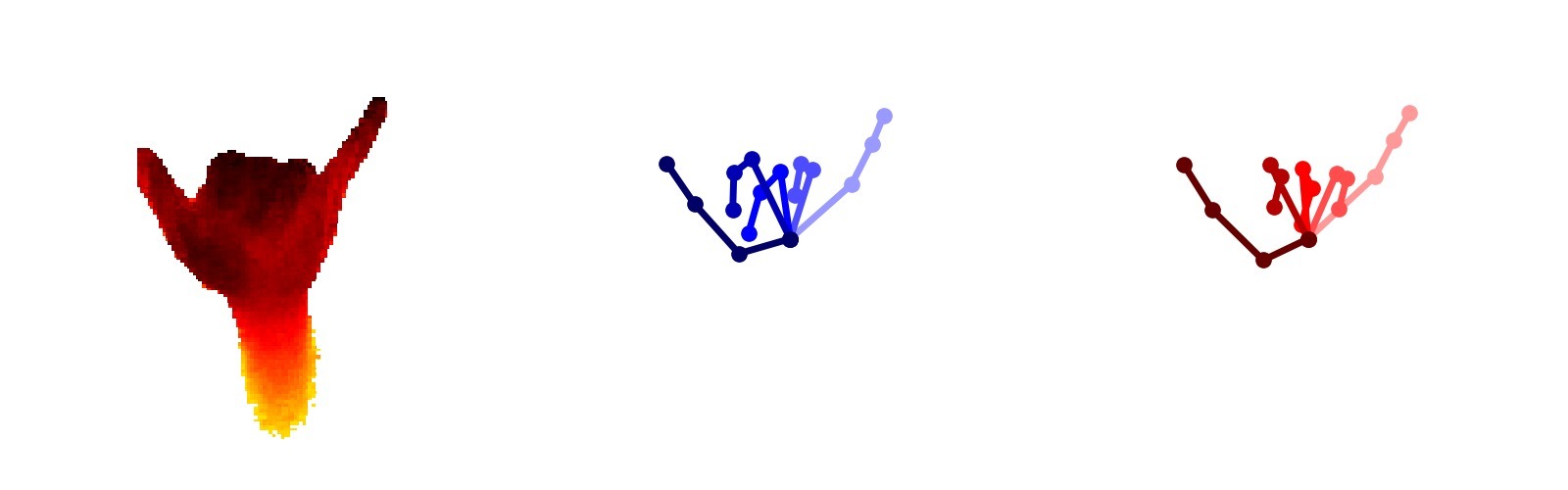}
\includegraphics[width=0.32\linewidth,clip,trim={0 18mm 0 8mm}]{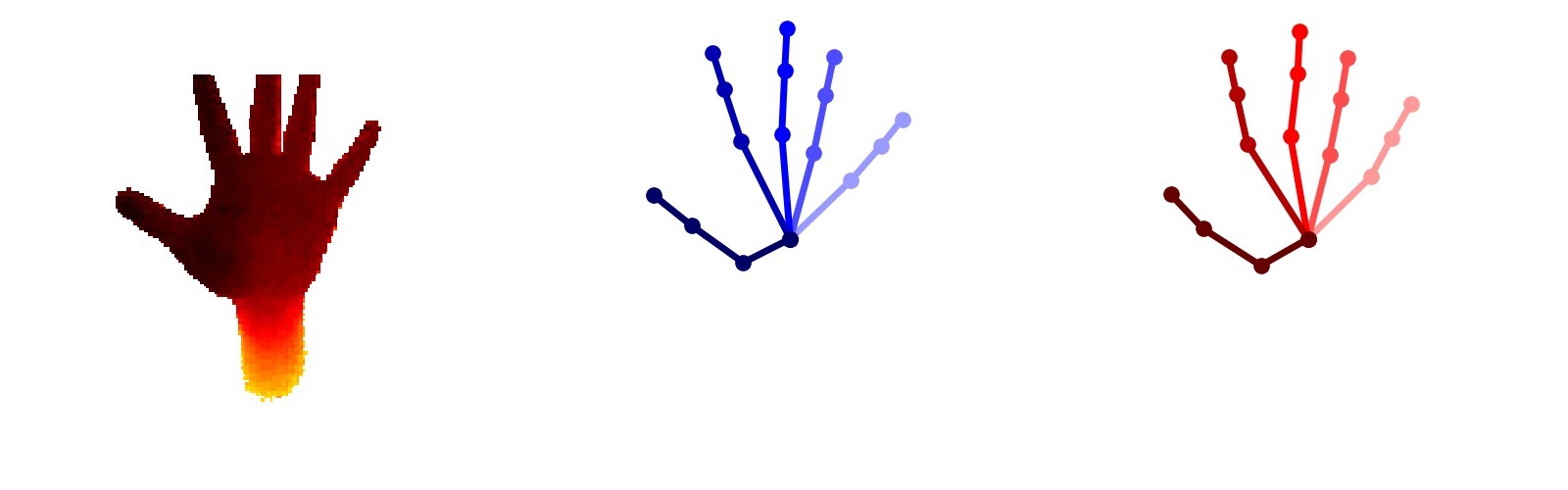}
\includegraphics[width=0.32\linewidth,clip,trim={0 18mm 0 8mm}]{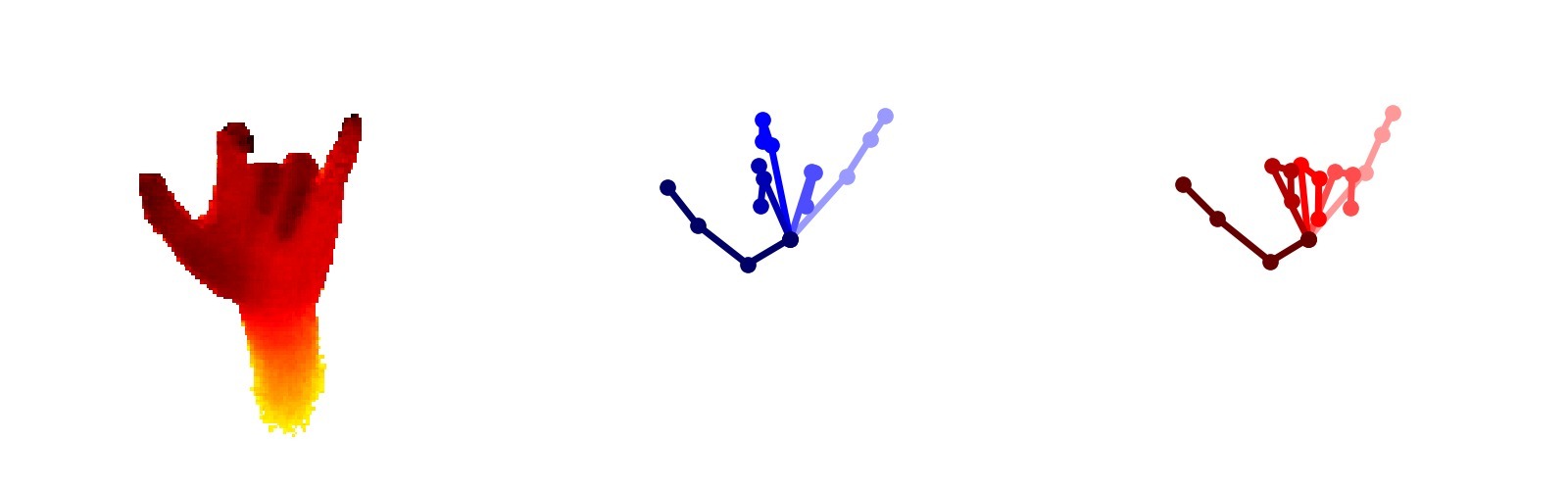}
\includegraphics[width=0.32\linewidth,clip,trim={0 18mm 0 8mm}]{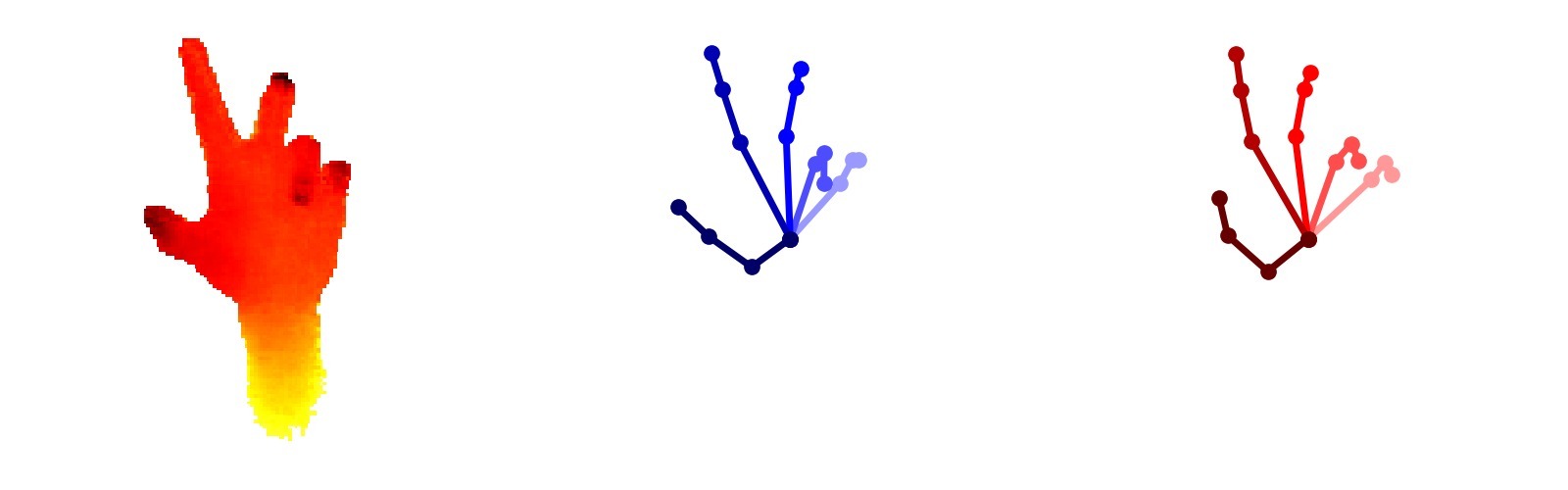}
\includegraphics[width=0.32\linewidth,clip,trim={0 18mm 0 8mm}]{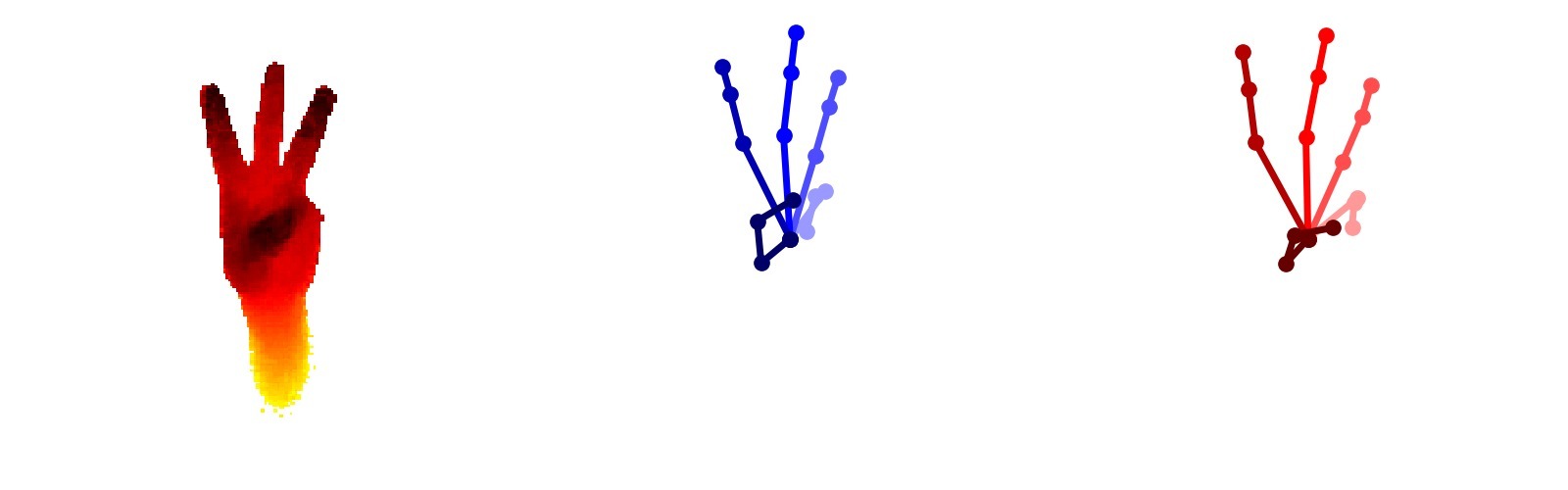}
\caption{\textbf{Depth to 3D joint predictions}. For each row-triplet, the left most column corresponds to the input image, the middle column is the ground truth 3D joint skeleton and the right column is our corresponding prediction.}
\label{fig:res_depth}
\end{figure*}

\begin{table}
\centering
\begin{tabular}{|l|c|c|c|}
\hline
 & \begin{tabular}{@{}c@{}}2D$\rightarrow$3D \\ RHD\end{tabular} & \begin{tabular}{@{}c@{}}RGB$\rightarrow$3D \\ RHD\end{tabular} & \begin{tabular}{@{}c@{}}RGB$\rightarrow$3D \\ STB\end{tabular} \\
\hline\hline
Variant 1 & 14.68 & \textbf{16.74} & 7.44\\
Variant 2 & 15.13 & 16.97 & 7.39\\
Variant 3 & \textbf{14.46} & 16.96 & \textbf{7.16}\\
Variant 4 & 14.83 & 17.30 & 8.16\\
\hline
\end{tabular}
\caption{The median end-point-error (EPE). Comparing our variants.}
\label{table:var_comparison}
\end{table}

\begin{table}
\centering
\begin{tabular}{|l|c|c|c|}
\hline
 & \begin{tabular}{@{}c@{}}2D$\rightarrow$3D \\ RHD\end{tabular} & \begin{tabular}{@{}c@{}}RGB$\rightarrow$3D \\ RHD\end{tabular} & \begin{tabular}{@{}c@{}}RGB$\rightarrow$3D \\ STB\end{tabular} \\
\hline\hline
\cite{zimmermann2017} (T+S+H) & 18.84 & 24.49 & 7.52\\ \hline
Ours (T+S+H) & \textbf{14.46} & \textbf{16.74} & \textbf{7.16}\\
Ours (T+S) & 14.91 & 16.93 & 9.11\\
Ours (T+H) & 16.41 & 18.99 & 8.33\\
Ours (T) & 16.92 & 19.10 & 7.78\\
\hline
\end{tabular}
\caption{The median end-point-error (EPE). Comparison to related work}
\label{table:rgb_prior_work_comparison}
\end{table}

\end{document}